# A Deep Multiscale Neural Network for Accurate Neurological Disorder Detection from MRI Scans and Real-Time Web Deployment

Ali Fatahi [1,2,*], Hoda Zamani [1,2,*], Mohammad H. Nadimi-Shahraki [3]

[1] Department of Computer Engineering, Na.C., Islamic Azad University, Najafabad, 8514143131, Iran
[2] Big Data Research Center, Na. C., Islamic Azad University, Najafabad, 8514143131, Iran
[3] International Graduate School of Artificial Intelligence, National Yunlin University of Science and Technology, Douliou, Yunlin 640301, Taiwan

* The article has two corresponding authors: Ali Fatahi, corresponding email: fatahi.ali.edu@gmail.com, ORCID: 0000-0002-7779-3470; and Hoda Zamani, corresponding email: hoda_zamani@yahoo.com, ORCID: 0000-0003-0444-4509.

**Abstract -** Neurological disorders involve diverse pathologies of the brain and nervous system, making early and accurate detection essential. While many deep CNNs have been developed for MRI-based classification of neurological disorders, most are optimized for binary tasks and often fail to capture the multi-class features needed to distinguish subtle anatomical differences across conditions. This study proposes the Enhanced Neurological Disorder Detection Network (End-Net) for multi-class MRI classification of neurological disorders. End-Net includes 24 convolutional layers, beginning with convolutional blocks followed by 21 optimized inception modules. These modules extract multiscale features via parallel 1×1, 3×3, and factorized 5×5 convolutional branches, along with max pooling, enabling the model to capture complementary texture, edge, shape, and contextual information. A global average pooling head, compact fully connected classifier, and dropout reduce parameters, limit overfitting, and improve robustness. End-Net was evaluated on the Multi-Class Neurological Disorder dataset, comprising MRI scans from patients with Alzheimer's disease, brain tumors, multiple sclerosis, and healthy controls. Severe class imbalance was addressed by augmenting minority classes with WGAN-GP and randomly undersampling the majority class. The results show that End-Net outperforms existing architectures in both accuracy and generalization. The model is also integrated into an online system for real-time web-based inference and accessibility.

**Keywords**: Deep Learning, Deep Convolutional Neural Networks, Inception module, Neurological Disorders, Magnetic Resonance Images data

## 1. Introduction

Neurological disorders, such as Alzheimer's disease, multiple sclerosis, and brain tumors, significantly impair cognitive and physical functions. [1]. These disorders can significantly affect normal bodily functions, leading to symptoms such as cognitive decline, motor difficulties, sensory issues, and emotional challenges [2]. Early diagnosis of these disorders is critical, as it can significantly raise the five-year survival rate to 75% [3]. However, the brain's intricate architecture makes detecting and treating neurological disorders particularly challenging, underscoring the importance of developing effective early-detection methods [4]. Deep learning (DL) algorithms are a subfield of AI and machine learning inspired by the structure and function of the human brain [5]. While they do not accurately replicate human cognition, DL models are highly effective at learning complex patterns from large datasets. They have demonstrated strong performance in analyzing neuroimaging data [6] by leveraging large datasets from modalities such as Single Photon Emission Computed Tomography (SPECT), Computed Tomography (CT), and Magnetic Resonance Imaging (MRI) to extract complex patterns and features that traditional methods often overlook. This enables more accurate diagnoses and deeper insights into neurological disorders. Common DL architectures for analyzing neurological images include Convolutional Neural Networks (CNNs) [7, 8], Long Short-Term Memory networks (LSTMs) [9], Recurrent Neural Networks (RNNs) [10], autoencoders [11], Generative Adversarial Networks (GANs) [12, 13], and transformers [14, 15]. CNNs are a compelling DL framework well-suited for image and video recognition tasks [5, 16, 17].

CNNs have demonstrated remarkable effectiveness in visual pattern recognition, particularly in processing medical imaging data such as MRI, CT, and SPECT scans. These models identify spatial hierarchies and patterns within images, making them ideal for object detection, image classification, and segmentation [18, 19]. Prominent CNN architectures like AlexNet [20], VGG16 [21], VGG19 [21], GoogLeNet [22], ResNet [23], U-Net [24], DenseNet [25], NeuroNet19 [26], and EfficientNet [27], NeuroNet19 [26], and C-DCNN [28] have been pivotal in advancing medical image analysis. The AlexNet model, proposed by Alex Krizhevsky et al. [20], comprises five convolutional layers and three fully connected layers, resulting in a total of eight learnable layers. It uses the Rectified Linear Unit (ReLU) activation function [29] and dropout to mitigate overfitting, and incorporates overlapping max pooling, which significantly enhances performance on large-scale image classification tasks [30]. The VGG16 model is a deep convolutional neural network architecture comprising 13 convolutional layers and 3 fully connected layers, totaling 16 weight layers, and uses 3×3 convolution filters with max pooling [31]. It uses ReLU activation and fully connected layers for classification. VGG19 builds upon VGG16 by adding three more convolutional layers, thereby improving its feature-extraction capacity while increasing computational cost [32]. Inception-V3 [33], an evolution of GoogLeNet [22], utilizes inception modules that apply multiple filter sizes in parallel and also employs techniques such as factorized convolutions and label smoothing to improve computational efficiency and accuracy.

Densely Connected Convolutional Networks (DenseNet) [25] enhance accuracy and efficiency by connecting each layer to all previous layers, thereby improving gradient flow and feature reuse. Transition layers are used between dense blocks to reduce the feature map size and control the number of parameters. Residual Networks (ResNet) [23] revolutionized DL by introducing residual connections, enabling the development of robust networks. Notable variants include ResNet-50, ResNet-101, and ResNet-152, which denote the number of layers in the network. NeuroNet19 [26], developed for brain tumor classification, incorporates an Inverted Pyramid Pooling Module to enhance multiscale feature extraction, thereby improving robustness and classification performance. A recent review by Zamani et al. [34] investigated deep CNN variants for neurological diagnosis. The authors revealed that research in this area is still largely driven by neuroimaging-focused CNNs, with relatively few studies evaluating multi-class performance or incorporating broader multimodal data. This highlights the need for more generalizable models. Hence, despite the success of the aforementioned deep CNN models and their variants across various domains, challenges remain in applying them to classify multiple neurological disorders from MRI scans [35-37]. Existing models are often tailored for binary classification, meaning they are designed to detect only one specific disease and its associated classes, such as differentiating between different stages of Alzheimer's disease. As a result, these models may not efficiently capture the multi-class features required to identify subtle anatomical changes across various conditions.

To address this limitation, the proposed Enhanced Neurological Disorder Detection Network (End-Net) introduces a novel multi-class classification approach that enables the detection and differentiation of multiple neurological disorders at various stages, including brain tumors, Alzheimer’s disease, and multiple sclerosis, within a uniform framework. The proposed End-Net comprises 24 convolutional layers, starting with a series of three-layer convolutional blocks, followed by 21 layers of optimized inception modules designed to capture multiscale features efficiently. These inception modules use parallel convolutional branches with 1×1, 3×3, and factorized 5×5 convolutions, along with max pooling. This multi-branch design empowers the model to capture both fine-grained and coarse-grained patterns simultaneously, thereby enhancing its ability to generalize across diverse datasets. Additionally, the model incorporates global average pooling and fully connected layers with dropout regularization, which help mitigate overfitting and improve robustness.

Moreover, a Multi-Class Neurological Disorder (MCND) dataset has been developed to assess the performance of the proposed model and competing models in large-scale multi-class neurological classification. The MCND dataset comprises MRI scans from individuals diagnosed with Alzheimer's disease, brain tumors, and multiple sclerosis, as well as scans from healthy control subjects. To address the severe class imbalance in the Alzheimer's subset, Wasserstein Generative Adversarial Networks with Gradient Penalty (WGAN-GP) is employed to oversample the two minority classes. At the same time, random undersampling is applied to the majority class, resulting in a balanced distribution of 1,500 samples per class. This approach stabilizes training, mitigates mode collapse, and enhances the model's ability to learn subtle, class-specific morphological features, critical for robust, generalizable classification. Two experiments were conducted to evaluate the proposed End-Net model against several well-known CNN architectures, including CNN, AlexNet [20], VGG19 [21], Inception-V3 [33], DenseNet-121 [25], NeuroNet19 [26], and C-DCNN [28]. The first assessed model performance on separate datasets for Alzheimer's disease, brain tumors, and multiple sclerosis, highlighting disorder-specific learning. The second tested generalization using the multi-disorder MCND dataset. The comparison results for accuracy, precision, recall, F1 score, cross-entropy loss, Mean Squared Error (MSE), Receiver Operating Characteristic (ROC) plot, and confusion matrix are analyzed. Moreover, visualization of End-Net's decision-making process using Gradient-weighted Class Activation Mapping (Grad-CAM) shows that the proposed model accurately identifies specific regions in MRI scans that significantly affect its predictions. Furthermore, the findings reveal that the proposed End-Net surpasses its competitors, attaining an accuracy of 0.9761, precision of 0.9766, recall of 0.9761, F1 score of 0.9763, cross-entropy loss of 0.1024, and MSE of 0.0050 on the MCND dataset. The key contributions of this paper are outlined below:

- Proposing the End-Net as a DL model to classify large volumes of MRI scans related to neurological disorders.

- The proposed End-Net features a 24-layer convolutional architecture, starting with three initial convolutional blocks and progressing through 21 optimized Inception modules, which facilitate effective multiscale feature extraction.
- Developing a multi-class neurological dataset named MCND, which combines MRI scans from publicly available datasets of neurological disorders, organized into eight classes covering various conditions, including brain tumors, Alzheimer's disease, and multiple sclerosis.
- Developing a real-time End-Net web-based platform to classify neurological disorders, accessible at https://end-net.aifatahi.space.
- The findings indicate that the proposed End-Net can proficiently classify neurological disorders from the MCND dataset, outperforming well-known contender models.

The rest of this paper is structured as follows: Section 2 reviews recent deep neural network architectures designed to detect neurological disorders from MRI scans. Section 3 presents the proposed End-Net architecture. Section 4 describes the development of a multi-class dataset for neurological disorders. Section 5 presents the experimental evaluations, followed by a discussion. Lastly, Section 6 concludes the paper and explores potential future research directions.

## 2. Related Works

This section reviews recent deep neural network architectures for detecting neurological disorders from MRI scans, including brain tumors, Alzheimer's disease, and multiple sclerosis.

**Deep neural networks to detect Alzheimer's disease from MRI scans:** Alzheimer's disease is a progressive neurodegenerative condition marked by the degeneration of brain cells, resulting in memory loss, cognitive decline, a reduction in essential skills, and, ultimately, death [38, 39]. Timely diagnosis is critical, as it enables early interventions that can slow disease progression and improve patients' quality of life [40]. Several DL models have recently been introduced to enhance the diagnostic process of Alzheimer's disease [41].

Janghel et al. [42] proposed a deep learning framework for analyzing medical imaging data by first converting 3D brain scans into 2D images and resizing them. This preprocessing step was followed by feature extraction using a CNN based on the VGG-16 architecture. The framework was applied to the ADNI dataset [43], which includes fMRI and PET images from patients with Alzheimer's disease and healthy controls. The model achieved 99.95% accuracy on the fMRI data and 73.46% on the PET data, demonstrating strong performance, particularly with fMRI. EL-Geneedy et al. [44] proposed a deep learning framework that uses a CNN with multiple convolutional and pooling layers, a class-weighted loss function to handle class imbalance, dropout regularization, and a softmax layer for multi-class classification of Alzheimer's disease stages from MRI data. The framework was tested on 6,400 brain MRIs, achieving 99.68% accuracy, 100% sensitivity, 100% specificity, and an AUC of 100%. However, since the model was trained and tested only on the Open Access Series of Imaging Studies (OASIS) [45], its generalizability to other populations or imaging settings may be limited.

Houria et al. [46] proposed a multimodal MRI fusion approach to detect white matter alterations and gray matter atrophy in patients with AD. The method integrates diffusion tensor imaging and structural MRI features, using a 2D CNN for feature extraction and a support vector machine for classification. The study utilized a balanced dataset comprising 150 subjects (50 with AD, 50 with CN, and 50 with MCI). It achieved classification accuracies of 99.79% for AD versus cognitively normal (CN), 99.6% for AD versus mild cognitive impairment (MCI), and 97.00% for MCI versus CN. However, the study is limited by its relatively small sample size and single-center validation, which may reduce the generalizability of the findings. Basaia et al. [47] developed a deep CNN aimed at diagnosing Alzheimer's disease and mild cognitive impairment (MCI) using individual structural MRI scans. The model was trained on 3D T1-weighted MR images and achieved up to 99% accuracy in distinguishing AD patients from healthy controls when applied to the Alzheimer's Disease Neuroimaging Initiative (ADNI) dataset [43]. A similar accuracy of 98% was obtained when the model was trained on a combined dataset from ADNI and the Milan dataset. However, its accuracy decreased to 75% when attempting to differentiate between MCI patients who later converted to AD and those who remained stable, even across both datasets.

Venugopalan et al. [48] proposed a deep learning framework that combines MRI, genetic, and clinical data to classify Alzheimer's disease. Using stacked denoising autoencoders, 3D CNNs, and an interpretation method based on clustering and perturbation analysis, the model identified key disease features and outperformed traditional methods. However, its reliance on the ADNI dataset may limit its generalizability to more diverse populations. Spasov et al. [49] developed a deep learning model with dual learning and 3D separable convolutions to predict MCI progression to Alzheimer's disease using MRI, clinical, and genetic data from 785 ADNI participants. Their multitask model achieved 86% accuracy and an AUC of 0.925 for MCI conversion prediction, and perfect classification between AD and controls, while using fewer parameters to reduce overfitting. However, the study employed single-center validation, which may limit the model's generalizability to more diverse populations. Shojaei et al. [50] developed a hybrid explainable AI approach for diagnosing Alzheimer's disease using MRI data from the ADNI dataset. Their approach combines a genetic algorithm-based Occlusion Map with backpropagation-based explainability techniques, applied to a 3D CNN. The approach initially used all 96 brain regions defined by the Harvard-Oxford atlas, achieving 96% validation accuracy. The approach identified and focused on 29 key brain regions to improve interpretability and performance, resulting in a refined model with 93% validation accuracy. When evaluated using 5-fold cross-validation, the model achieved an overall classification accuracy of 87%.

Ayus et al. [51] proposed two hybrid deep learning models, CNN-Conv1D-LSTM and the Hybrid Recurrent Ensemble Network (HReENet), for the automated detection of Alzheimer's disease. The CNN-Conv1D-LSTM integrates a Conv1D-LSTM classifier with convolutional neural networks for feature extraction, while HReENet ensembles the outputs of the CNN, LSTM, and CNN-Conv1D-LSTM to improve robustness. Using 5-fold cross-validation, the CNN-Conv1D-LSTM achieved 98.75% accuracy, while the HReENet reached 99.97% accuracy, with similarly high precision, recall, and F1-scores. The study also introduced ALZ-IS, an online diagnostic interface based on HReENet. A recent study addresses the scarcity and severe class imbalance of the Kaggle Alzheimer's MRI dataset by synthesizing class-conditional images using WGAN-GP. The generated MRIs showed near-realistic quality (mean FID≈0.13, SSIM≈0.97, PSNR≈32 dB) and, when used to balance the training set, outperformed conventional oversampling, SMOTE/ADASYN, augmentation, and class weighting. On a fixed test split, adding synthetic data improved balanced accuracy by 11.77% and MCC by 15%, with ~91.4% gains on minority classes at a cost of only ~1% on the majority class [38]. Complementing these findings, Yuda et al. [52] built a WGAN-GP–centric pipeline on the same Kaggle dataset to synthesize minority MRIs and rebalance the training. Beyond classification, it quantified realism and analyzed separability with GLCM features and 2D/3D UMAP. Using a Random Forest, accuracy on a held-out set rose from 30.46% (imbalanced) to 56.84% after augmentation, with tighter UMAP clusters and confusion maps evidencing improved class structure. Various deep learning models have been introduced for Alzheimer's disease diagnosis, such as AlzheimerNet [53], ADD-Net [54], DEMNET [55], OViTAD [56], Fuzzy-VGG [57], Xception [58], VGG19 [59], and FDN-ADNet [60], each employing different network architectures and techniques.

**Deep neural networks for brain tumor detection from MRI scans:** Early identification of brain tumors is crucial, and deep learning algorithms show significant promise in enhancing the accuracy of tumor detection and classification for healthcare providers [61]. Haq et al. [62] introduced a multi-level convolutional neural network (MCNN) model to classify brain tumor types—Meningioma, Glioma, and Pituitary tumors—using MRI brain imaging data. The model, enhanced with data augmentation and transfer learning, achieved 99.89% accuracy on a hold-out validation set. Saeedi et al. [63] proposed two deep learning architectures for brain tumor classification: a custom 2D CNN and a convolutional autoencoder-based network. These models were designed to differentiate among glioma, meningioma, and pituitary tumors. The 2D CNN achieved a training accuracy of 96.47%, while the autoencoder-based model reached 95.63%. Furthermore, the reported results indicate that the average recall was 95% for the 2D CNN and 94% for the convolutional autoencoder-based network. Majib et al. [64] presented a VGG Stacked Classifier Network (VGG-SCNet) for detecting brain tumors in MRI scans. VGG-SCNet integrates various traditional and hybrid machine learning models, meticulously constructed and analyzed to autonomously classify brain tumor images. The VGG-SCNet achieved precision, recall, and F1 scores of 99.2%, 99.1%, and 99.2%, respectively.

Anantharajan et al. [65] introduced a DL-based method for brain tumor detection in MRI scans. This approach involves preprocessing the images, using fuzzy c-means segmentation, extracting features from the grey-level co-occurrence matrix, and employing an ensemble deep neural network support vector machine to classify abnormal tissues. The method achieved 97.93% accuracy, 92% sensitivity, and 98% specificity in brain tumor detection. Aamir et al. [66] proposed an automated brain tumor detection method to address the limitations of manual diagnosis, including reliance on radiologist expertise, inefficiency, and error-proneness. The approach begins by enhancing the visual quality of brain MR images through preprocessing. Then, features are extracted using two pre-trained deep learning models and fused into a hybrid feature vector via partial least squares. Tumor locations are identified using agglomerative clustering, and the resulting proposals are resized before being classified by the network. This method achieved a classification accuracy of 98.95%. However, it was tested on a single-center validation dataset, which may limit its generalizability across diverse clinical environments and imaging conditions.

Ottom et al. [67] introduced the Znet architecture for 2D brain tumor segmentation in MRI. Znet utilizes skip connections and encoder-decoder architectures, combined with data augmentation techniques, to generalize from a small number of expert-delineated tumors. However, since it was evaluated using a single-center validation dataset, its applicability may be limited across different clinical settings and imaging protocols. Haque et al. [26] introduced NeuroNet19 to classify brain tumors from MRI data. NeuroNet19 employs the VGG19 framework and features a distinctive inverted pyramid pooling module. This module captures multiscale feature maps to effectively extract both local and global contexts, thereby improving feature map quality regardless of tumor size or location. Furthermore, NeuroNet19 integrates explainable AI via local interpretable model-agnostic explanations, generating saliency maps that highlight clinically relevant neuroimaging features contributing to predictions, thereby increasing transparency and accountability. Amin et al. [68] introduced a deep neural network architecture comprising three convolutional layers, three ReLU activation layers, and a softmax activation function. The deep NN assigns labels based on these center pixels to perform segmentation. Onakpojeruo et al. [69] present a Pix2Pix GAN-based framework for generating synthetic MRI data to support brain tumor classification, addressing dataset scarcity and privacy challenges. They introduce a conditional deep convolutional neural network (C-DCNN) to classify glioma, meningioma, pituitary, and healthy cases, achieving 86% accuracy on synthetic data. The C-DCNN outperformed models like ResNet50, VGG16/19, and InceptionV3. The deep NN assigns labels based on these center pixels to perform segmentation. Other variants for detecting brain tumors from MRI include TumorDetNet [70], Brainnet [71], YOLOv8 [72], Hybrid-NET [73], JGate-AttResUNet [74], an XAI-enhanced efficientNetB0 [75], DeepSeg [76], and enhanced TumorNet [77].

**Deep neural networks to detect Multiple Sclerosis (MS) from MRI scans:** Recent developments in DL algorithms show potential for detecting early symptoms of MS, which may facilitate earlier diagnosis and treatment and decrease overall costs [78, 79]. Acar et al. [80] developed an efficient CNN model with a small number of trainable parameters to identify Multiple Sclerosis lesions from FLAIR MR images. When tested using 5-fold cross-validation, the model achieved up to 98% accuracy at the slice level and over 90% at the patient level. Yoo et al. [81] applied a DL method to identify MS pathology in MRI scans of normal-appearing brain tissue. The method employs unsupervised DL, combined with random forest algorithms, applied to myelin and T1-weighted images. Tested on 55 MS patients and 44 healthy controls with 11-fold cross-validation, their approach achieved 87.9% accuracy, outperforming traditional regional measurements and single-modality deep learning features. However, the study was conducted at a single center, which may limit the generalizability of the findings. Eitel et al. [82] introduced a transparent deep learning framework for diagnosing multiple sclerosis, leveraging 3D CNNs integrated with layer-wise relevance propagation. The model was initially pre-trained on a large ADNI dataset and subsequently fine-tuned to differentiate between MS patients ($n = 76$) and healthy controls (n = 71). It achieved a balanced accuracy of 87.04% with an area under the ROC curve of 96.08%. Acar et al. [80] introduced a lightweight CNN model for detecting multiple sclerosis lesions in brain fluid-attenuated inversion recovery MRI scans, emphasizing low computational complexity. However, single-center validation limits the model's generalizability to broader, multi-center clinical settings.

Kamraoui et al. [83] proposed the DeepLesionBrain model for lesion segmentation in multiple sclerosis. The DeepLesionBrain model uses an ensemble of compact 3D CNNs with extensive overlapping receptive fields. This spatially distributed architecture enhances robustness by mitigating the risk of generalization failure among individual networks. In addition, DeepLesionBrain incorporates an image quality-based data augmentation technique to reduce reliance on specific acquisition protocols. It also employs hierarchical specialization learning, in which a generic network is pre-trained on whole-brain data, and its weights are transferred to specialized subnetworks for localized lesion analysis. The model was tested across multiple datasets, including MSSEG'16, the ISBI Challenge, and in-house clinical scans. Musall et al. [84] proposed a Diffusely Abnormal White Matter Network (DAWM-Net) that utilizes semi-supervised deep learning to detect MS-related abnormalities in multiparametric brain MRI scans. The framework focuses on the automated segmentation of diffusely abnormal white matter (DAWM), an underexplored yet clinically relevant MRI feature in MS. By leveraging limited labeled data, DAWM-Net outperforms traditional intensity thresholding methods, enabling more accurate and scalable analysis of DAWM to characterize early or distinct MS pathology.

Pandian et al. [85] developed a Hybrid Deep CNN (HD-CNN) framework for classifying Multiple Sclerosis using MRI slices. The method combines chaotic leader-selective particle swarm optimization with a refined slime mold algorithm to extract hidden white-matter features from pixel-intensity ratios. Then it selects the most relevant features for segmentation. An embedded maximum-bidirectional-gradient classifier is then used to detect MS lesions by identifying white matter abnormalities. The HD-CNN was evaluated on the UMCL [86] and ISBI 2015 [87] benchmark datasets, achieving classification accuracies of 98.56% on FLAIR, 93.15% on T1W, and 98.44% on T2W images. Models such as Deep attention V-Net [88], CVSnet [89], RimNet [90], modified attention U-Net [91], DeepCONN [92], SynergyNet [93], and Recurrent Slice-wise Attention Network (RSANet) [94] are among the deep learning approaches developed for MS detection.

## 3. Enhanced Neurological Disorder Detection Network (End-Net)

The Enhanced Neurological Disorder Detection Network (End-Net) is designed for multi-class classification of neurological disorders using MRI scans. The network architecture integrates convolutional blocks, optimized inception modules, global average pooling, and fully connected layers, enabling it to capture both local anatomical details and global contextual features. Let the input MRI scan be denoted as: $X_0 \in R^{H \times W \times C}$, where $H$ (Height) is the number of vertical pixels, $W$ (Width) is the number of horizontal pixels, and $C$ (Channels) is the number of color channels.

### 3.1 Convolutional Blocks

The convolutional layers extract spatial features from MRI scans by sliding filters over the input image. The convolution operation is defined using Eq. (1), where $F_l^{(f)}(i,j)$ is output at position $(i,j)$ of filter $f$ in layer $l$, $W_l^{(f)}(u,v,c)$ is the weight of filter $f$ at kernel position ($u$, $v$) for channel $c$, $b_l^{(f)}$ is a bias term, $K$ is the kernel size, and $C_l - 1$ is the number of channels in the previous layer.

$$F_l^{(f)}(i,j) = \sum_{u=0}^{K-1} \sum_{v=0}^{K-1} \sum_{c=1}^{C_{l-1}} W_l^{(f)}(u,v,c)\, X_{l-1}(i+u, j+v, c) + b_l^{(f)} \tag{1}$$

The model's input consists of MR images preprocessed to a standard resolution of (210, 210, 3) to ensure uniformity across the dataset. Each convolutional block begins with a 3×3 convolution operation, followed by a ReLU activation layer that introduces non-linear transformations, as denoted in Eq. (2).

$$A_l^{(f)}(i,j) = \max\left(0, F_l^{(f)}(i,j)\right) \tag{2}$$

A batch normalization layer is applied after each 3×3 convolution in the three main convolutional blocks (32, 64, and 128 filters), and before the subsequent 2×2 max-pooling operation to maintain activation stability and accelerate model convergence during training, as defined using Eq. (3). In this equation $\mu_l^{(f)}$ and $\sigma_l^{(f)}$ are the mean and variance of activations computed over the mini-batch, $\gamma_l^{(f)}$ and $\beta_l^{(f)}$ relearnable scale and shift parameters and $\epsilon$ is a small constant for numerical stability.

$$\hat{A}_l^{(f)}(i,j) = \gamma_l^{(f)} \frac{A_l^{(f)}(i,j) - \mu_l^{(f)}}{\sqrt{\sigma_l^{(f)} + \epsilon}} + \beta_l^{(f)} \tag{3}$$

This combination enables the network to extract core features, including edges, textures, and gradients, which are crucial for differentiating various stages of neurological disorders. After each convolutional block, a max-pooling layer downscales the spatial dimensions of the feature maps, thereby reducing computational overhead while preserving key features, as shown in Eq. (4).

$$X_l^{(f)}(i,j) = \max_{\substack{0\le m<K_p \\ 0\le n<K_p}} \hat{A}_l^{(f)}(s\,i+m, s\,j+n) \tag{4}$$

This combination enables the network to extract core features such as edges, textures, and gradients, which are crucial for differentiating stages of neurological disorders. After each convolutional block, a max-pooling layer downscales the spatial dimensions of the feature maps, thereby reducing computational overhead while preserving key features. This operation is shown in Eq. (4), where $K_p$ is the pooling size (e.g., 2 for 2×2 pooling), $s$ is the stride, $\hat{A}_l^{(f)}$ is the input activation, and $X_l^{(f)}$ is the pooled output. This formulation provides flexibility for different pooling sizes and strides, making it suitable for a wide range of network configurations.

### 3.2 Optimized Inception Modules

Conventional Inception modules, as introduced in early deep learning architectures such as GoogLeNet, utilize parallel branches of 1×1, 3×3, and 5×5 convolutions, along with 3×3 max pooling, to capture features at multiple scales. These modules often incorporate 1×1 convolutions for dimensionality reduction before the more computationally expensive 3×3 and 5×5 operations, and concatenate the outputs from each branch to enhance representational diversity. Later versions, such as Inception v2 and v3, further improved efficiency by replacing 5×5 convolutions with two stacked 3×3 convolutions. Building upon these ideas, the proposed optimized inception modules introduce several enhancements to better suit the complexity of MRI-based classification of neurological disorders. Each module is integrated after key convolutional stages and consists of parallel branches including 1×1 and 3×3 convolutions, a factorized 5×5 path (two sequential 3×3 convolutions), and a max-pooling branch followed by a 1×1 convolution. To improve adaptability and performance, the design incorporates three key modifications: (1) replacement of 5×5 convolutions with two 3×3 layers, reducing parameters while increasing non-linearity and model expressiveness; (2) uniform application of 1×1 convolutions across all branches, including post-pooling, to enhance efficiency and retain feature richness; and (3) adaptive filter scaling in each branch based on the depth of the preceding convolutional block, providing flexible control over receptive field size.

These enhancements aim to enhance sensitivity to subtle anatomical variations in brain MRI scans, thereby increasing the network's effectiveness in detecting fine-grained patterns of neurological disorders. The optimized Inception module processes feature maps through the following branches:

**Branch 1: 1×1 Convolutions**: Applied directly to the input to capture fine-grained features and reduce dimensionality, which helps preserve important information while minimizing computational burden. The operation is defined using Eq. (5), where $W_{1\times1}^{(f_1)}$ is the 1×1 convolution kernel of filter $f_1$, $b_{1\times1}^{(f_1)}$ is the bias term, and $f_1 = 1, 2, \ldots, F_1$, with $F_1$ being the number of the filters in Branch 1.

$$B_1^{(f_1)}(i,j) = \sum_{c=1}^{C} W_{1\times1}^{(f_1)}(c).X(i,j,c) + b_{1\times1}^{(f_1)} \tag{5}$$

**Branch 2: 1×1 Convolution followed by a 3×3 Convolution (Factorized 3×3 branch)**: This branch captures medium-scale patterns essential for detecting moderate structural abnormalities in the brain. The initial 1×1 convolution reduces dimensionality before applying the 3×3 convolution, improving computational efficiency. The operation is denoted in Eq. (6), where C′ is the reduced number of channels after the 1×1 convolution and $Z_2 = Conv_{1\times1}(X)$.

$$B_2^{(f_2)}(i,j) = \sum_{c=0}^{\acute{C}-1}\sum_{u=0}^{2}\sum_{v=0}^{2} W_{3\times3}^{(f_2)}(u,v,c)\, Z_2(i+u,j+v,c) + b_{3\times3}^{(f_2)} \tag{6}$$

**Branch 3: 1×1 Convolution followed by two 3×3 Convolutions (Factorized 5×5 branch)**: Instead of using a computationally expensive 5×5 convolution, this method factorizes it into two sequential 3×3 convolutions. This factorization reduces the number of parameters and computational costs while allowing the network to learn more complex features by adding non-linearities. This operation is presented in Eq. (7), where C″ denotes the number of channels after the first 3×3 convolution, $Z_3 = Conv_{1\times1}(X)$ and $\acute{Z}_3 = Conv_{3\times3}(Z_3)$.

$$B_3^{(f_3)}(i,j) = \sum_{c=0}^{C''-1}\sum_{u=0}^{2}\sum_{v=0}^{2} W_{3\times3}^{(f_3)}(u,v,c)\, \acute{Z}_3(i+u,j+v,c) + b_{3\times3}^{(f_3)} \tag{7}$$

**Branch 4: 3×3 Max Pooling followed by a 1×1 Convolution:** This branch captures localized contextual information by applying a 3×3 max pooling operation, which aggregates the most salient responses within a local neighborhood. The pooled features are then passed through a 1×1 convolution to introduce additional non-linearity and enable channel-wise feature transformation. When applied with a stride greater than one, the pooling operation also reduces spatial resolution. This operation is presented in Eqs. (8) and (9).

$$Z_4(i,j,c) = \max_{0 \le m,\ n<3} X(i+m, j+n, c) \tag{8}$$

$$B_4^{(f_4)}(i,j) = \sum_{c=0}^{C-1} W_{1\times 1}^{(f_4)}(c)\, Z_4(i,j,c) + b_{1\times 1}^{(f_4)} \tag{9}$$

The operations' outputs are concatenated along the depth axis, allowing the model to learn fine-grained details and larger contextual patterns simultaneously. This ability to combine multiple receptive fields at different scales is particularly beneficial in MRI classification, where neurological abnormalities can vary widely in size and shape across various stages of the disorder. After each inception module (except the final one), a max pooling layer is applied, reducing the spatial dimensions and focusing on the most salient features. This sequence of convolutional blocks, followed by inception modules and pooling layers, enables the network to progressively capture and condense the salient features required for accurate classification.

### 3.3 Fully Connected Layers and Output

After the final inception module, the model uses a global average pooling layer instead of flattening the feature maps, which helps retain spatial information while reducing them to a single vector for classification. The global average pooling layer is denoted using Eq. (10), where D is the total depth of the concatenated feature maps.

$$z_d = \frac{1}{H\,W} \sum_{i=0}^{H-1} \sum_{j=0}^{W-1} Y_{i,j,d}, \quad d = 0, 1, \ldots, D-1 \tag{10}$$

This pooling operation averages each feature map into a single value, significantly reducing the number of parameters and mitigating overfitting while preserving crucial features. The network comprises two fully connected layers with 512 and 256 neurons, each activated by ReLU, enabling it to extract high-level, abstract representations from neuroimaging data. To prevent overfitting and improve generalization, dropout layers with a 30% rate are applied after each dense layer as presented in Eq. (11), where $z$ is the feature vector obtained from global average pooling, $W_1$ and $W_2$ are weight matrices of the fully connected layers, $b_1$ and $b_2$ are bias vectors, $h_1$ and $h_2$ are hidden layer activations, and $p_d$ is a dropout probability.

$$
\begin{aligned}
h_1 &= ReLU(W_1 z + b_1) \\
h_1^{drop} &= Dropout(h_1, p_d) \\
h_2 &= ReLU\left(W_2 h_1^{drop} + b_2\right) \\
h_2^{drop} &= Dropout(h_2, p_d)
\end{aligned} \tag{11}
$$

The final output layer consists of eight neurons, each representing a class, and utilizes a softmax activation function to provide normalized probability estimates for each class, as denoted in Eq. (12), where $\hat{y}_c$ is the predicted probability for class *c*.

$$
\hat{y}_c = \frac{\exp\left(\left(W_0 h_2^{drop} + b_0\right)_c\right)}{\sum_{k=0}^{C-1} \exp\left(\left(W_0 h_2^{drop} + b_0\right)_k\right)}, \qquad c = 1, \ldots, C-1 \tag{12}
$$

This structure enables the model to classify input MRI scans based on the class with the highest predicted probability. By integrating global average pooling and optimizing fully connected layers with dropout, the model reduces complexity while maintaining high performance, ensuring accurate classification across all classes. The pseudocode and architecture of End-Net are presented in Algorithm 1 and Fig. 1, respectively. End-Net begins with an input layer that processes images standardized to 210 × 210 pixels with 3 RGB color channels. It includes 24 convolutional layers, starting with initial convolutional blocks consisting of three layers, followed by 21 layers of optimized inception modules for efficient multiscale feature extraction. The architecture also incorporates pooling and fully connected layers to support hierarchical representation learning. These are followed by convolutional blocks, which contain convolutional layers for feature extraction, batch normalization to enhance learning stability, and max pooling layers to reduce spatial dimensions. Inception modules are incorporated after these initial blocks, utilizing parallel convolutional paths with different kernel sizes to derive distinct features from the input image, thereby enhancing the model's ability to capture multiscale characteristics. After passing through the convolutional and inception layers, a global average pooling layer consolidates the feature maps into a single vector, simplifying the model and enhancing its robustness. Subsequently, several fully connected layers are used to analyze the extracted features and prepare them for classification. Finally, the network terminates with an output layer that utilizes softmax activation to produce normalized probability distributions over the target classes, enabling the model to perform effective multi-class classification.

**Algorithm 1.** Pseudocode of the proposed End-Net

---

**Input:** MCND dataset, Target label, learning rate, and decay factor
**Output:** Neurological disorders detection from the MCND dataset

1. Initialize learning rate and decay factor.
2. **Dataset Preparation:**
   2.1 Load MCND images.
   2.2 Resize to 210 × 210 × 3; normalize pixel intensities.
   2.3 Split into training, validation, and test sets.
3. **Define Inception Module:**
   3.1 Branch1: 1×1 convolution
   3.2 Branch2: 1×1 convolution → 3×3 convolution
   3.3 Branch3: 1×1 convolution → two successive 3×3 convolutions
   3.4 Branch4: 3×3 max pooling → 1×1 convolution
   3.5 Concatenate all branches; return output tensor
4. **Model Architecture:**
   4.1 Input layer: Shape = (210, 210, 3).
   4.2 For filters ∈ {32, 64, 128}:
      4.2.1 Convolution (3×3, filters) → BatchNorm → MaxPool (2×2)
      4.2.2 Inception Module(filters) → MaxPool (2×2)
   4.3 Global Average Pooling.
   4.4 Dense (512 units) → Dropout (0.3).
   4.5 Dense (256 units) → Dropout (0.3).
   4.6 Output layer: Dense layer with softmax activation.
5. **End-Net compilation**
6. **End-Net training on the training set with validation monitoring.**
7. **Validate End-Net with test data.**
8. **Measure and return performance results of the End-Net**

---

Algorithm 1 outlines the sequential stages for configuring, training, and evaluating the proposed End-Net. Step 1 establishes the learning rate and decay factor to regulate the model's learning dynamics. In Step 2, the dataset undergoes preparation through loading, resizing (210 × 210 pixels), and normalization to standardize inputs, followed by partitioning into training, validation, and test subsets. Step 3 defines the Inception Module, comprising four parallel branches with distinct convolutional and pooling operations that collectively enhance the model's ability to discern diverse image features across multiple scales. Step 4 details the network's architecture, incorporating convolutional layers, batch normalization, max pooling, and Inception Modules, culminating in global average pooling to consolidate features. It also includes fully connected layers with dropout to mitigate overfitting, as well as a softmax-activated output layer for multi-class image classification. In Step 5, the model is compiled by specifying the loss function, optimizer, and evaluation metrics. Step 6 entails training the model on the training set while monitoring its validation performance to confirm effective learning. Step 7 evaluates the trained model on an independent test set to assess its practical efficacy. Finally, Step 8 computes and reports performance metrics, including accuracy, recall, F1 score, precision, cross-entropy loss, and MSE, to quantify the model's effectiveness at identifying neurological disorders.

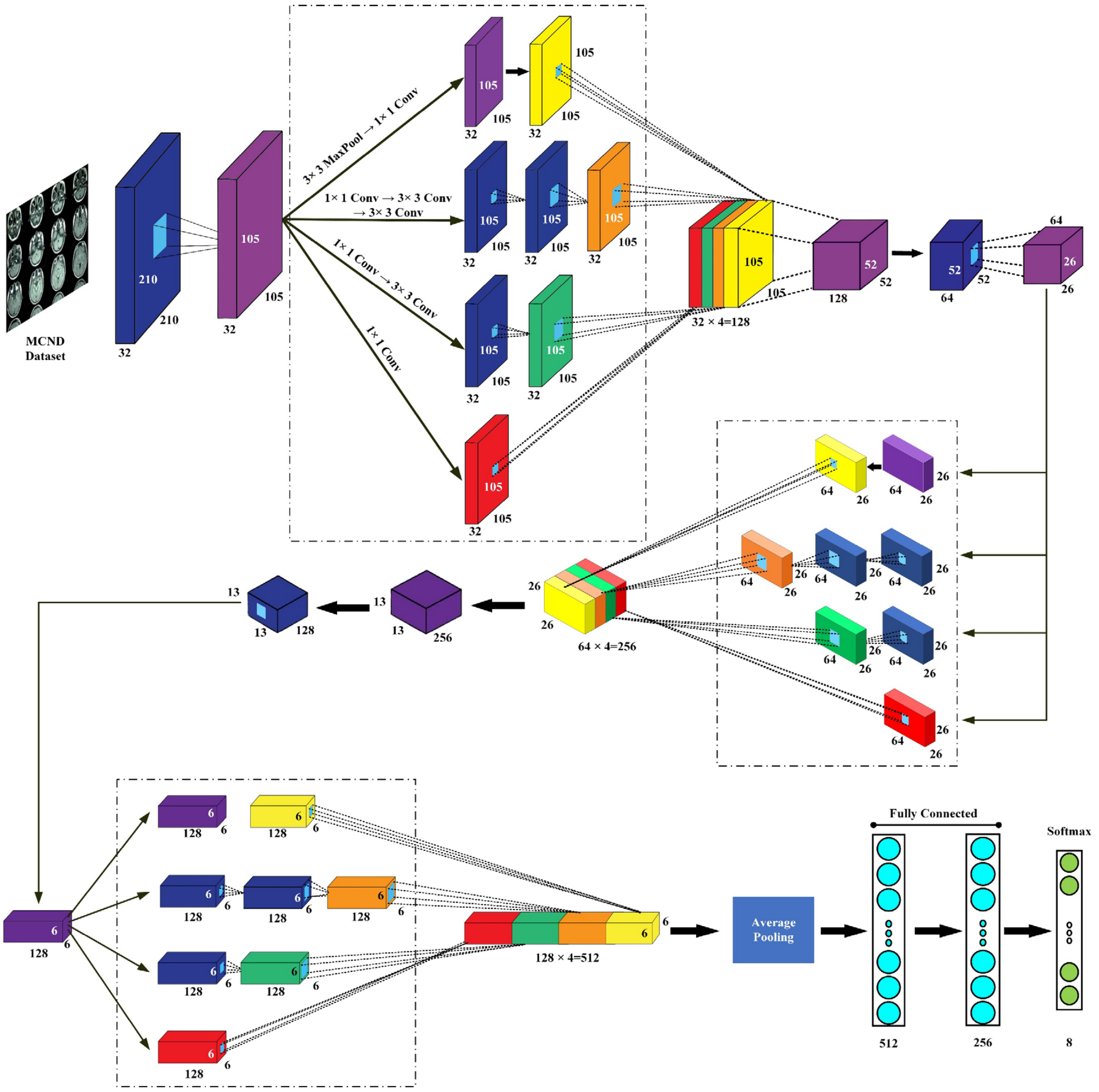

**Fig. 1** The architecture of the proposed End-Net

## 4. A Multi-Class Neurological Disorder Dataset

This section describes the Multi-Class Neurological Disorder (MCND) dataset, which integrates publicly available datasets from multiple neurological disorders for comprehensive evaluation. The MCND dataset comprises MRI data from three neurological conditions, available in the Kaggle repository: Alzheimer's disease (AD) [95], Brain Tumor (BT) [96], and Multiple

Sclerosis (MS) [97]. In addition, MRI scans from healthy subjects are sourced from the publicly available Information eXtraction from Images (IXI) dataset [98], which provides a diverse collection of brain MRI scans from healthy volunteers.

**Alzheimer's Disease (AD) MRI Dataset:** This dataset consists of 3,200 axial MRI scan samples from patients diagnosed with Alzheimer's disease, divided into three groups according to the stage of cognitive impairment: AD-Mild Dementia (AD-MD): 896 samples, AD-Moderate Dementia (AD-MoD): 64 samples, and AD-Very Mild Dementia (AD-VMD): 2,240 samples.

**Brain Tumor (BT) MRI Datasets:** The BT MRI dataset is compiled and published by Masoud Nickparvar on Kaggle [96]. It merges three distinct sources: the Figshare dataset [99], the SARTAJ dataset [100], and the Br35H dataset [101]. The combined dataset comprises 5,022 grayscale MRI scans diagnosed with disease, in JPEG format, categorized into three classes: BT-glioma (1,621 images), BT-meningioma (1,645 images), and BT-pituitary tumor (1,757 images).

**Multiple Sclerosis (MS) MRI Dataset:** The MS MRI dataset comprises axial and sagittal FLAIR MRI brain images acquired from 72 patients diagnosed with multiple sclerosis, a chronic autoimmune disorder affecting the central nervous system [97]. The dataset contains 1,441 brain images representing a diverse cohort of MS patients.

**Information eXtraction from Images (IXI) Dataset**: The Information eXtraction from Images (IXI) dataset [98] comprises nearly 600 MRI scans of healthy adult subjects. These scans were acquired at three hospitals in London using 1.5T and 3T MRI scanners, encompassing T1-, T2-, and PD-weighted images, as well as MRA and diffusion-weighted images (15 directions). In this study, we have used 581 T1- and 578 T2-weighted images.

### 4.1 Preprocessing Phase in the MCND Dataset

The MCND dataset was systematically assembled and preprocessed to ensure data quality and consistency, thereby facilitating effective model training and development. During dataset aggregation, 465 duplicate scans were identified and removed to improve accuracy, consistency, and reliability. Furthermore, MR images sourced from various institutions were carefully cropped and standardized to harmonize image formats and quality, resulting in a uniform, high-quality dataset suitable for robust analysis. The IXI dataset [98] originally consists of volumetric medical imaging scans stored in NIfTI format. Each 3D scan is iteratively processed to extract a representative 2D slice. Specifically, the algorithm identifies the middle slice in the axial plane, applies a 90° left rotation to ensure consistent orientation, and normalizes voxel intensities to [0, 255]. This ensures comparability across scans with different dynamic ranges. Finally, the preprocessed slices are converted into 8-bit grayscale PNG images and stored, producing a standardized 2D representation of each volumetric scan.

Figure 2 illustrates the class distribution of the MCND dataset before and after addressing imbalance using the Wasserstein GAN with Gradient Penalty (WGAN-GP) [38] and random undersampling techniques. The dataset is heavily imbalanced, with only 64 samples in AD-MoD, 896 in AD-MD, and 2,240 in AD-VMD, which risks biasing the model toward the majority class. To address this, WGAN-GP is used to augment AD-MoD and AD-MD, while random undersampling is used to reduce AD-VMD. This balances all classes in the Alzheimer's disease dataset to 1,500 samples each, yielding a uniform distribution for more robust, generalizable model training. Guided by these constraints, we adopt WGAN-GP as the oversampling engine because its Wasserstein objective and gradient-penalty–enforced 1-Lipschitz critic yield stable training under small sample regimes and mitigate mode collapse, which is crucial for capturing the subtle, class-specific morphology of minority Alzheimer's MRIs, while producing genuinely new examples rather than simple interpolations (e.g., SMOTE) or geometric variants (classical augmentation). Specifically, images are resized to 128×128 and scaled to the range of [-1, 1]; separate class-conditional WGAN-GP models are trained for AD-MoD and AD-MD. Each model uses a generator with batch normalization and a tanh output, and a critic with LeakyReLU activations [102], both trained with a gradient penalty term. Optimization follows Adam's standard practice of multiple critic updates per generator step. All architectural choices and hyperparameters (gradient-penalty coefficient, learning rate, $\beta$-coefficients, batch size, and latent dimensionality) are set to the optimal configuration for this Alzheimer's dataset, as tested and discussed in [38]; the exact values are presented in Table 1.

Throughout training, we monitor fidelity/diversity metrics and select the best checkpoint. From the chosen generator, we sample latent vectors to synthesize the required number of images (aiming for 1,500 per class) and integrate them into the dataset.

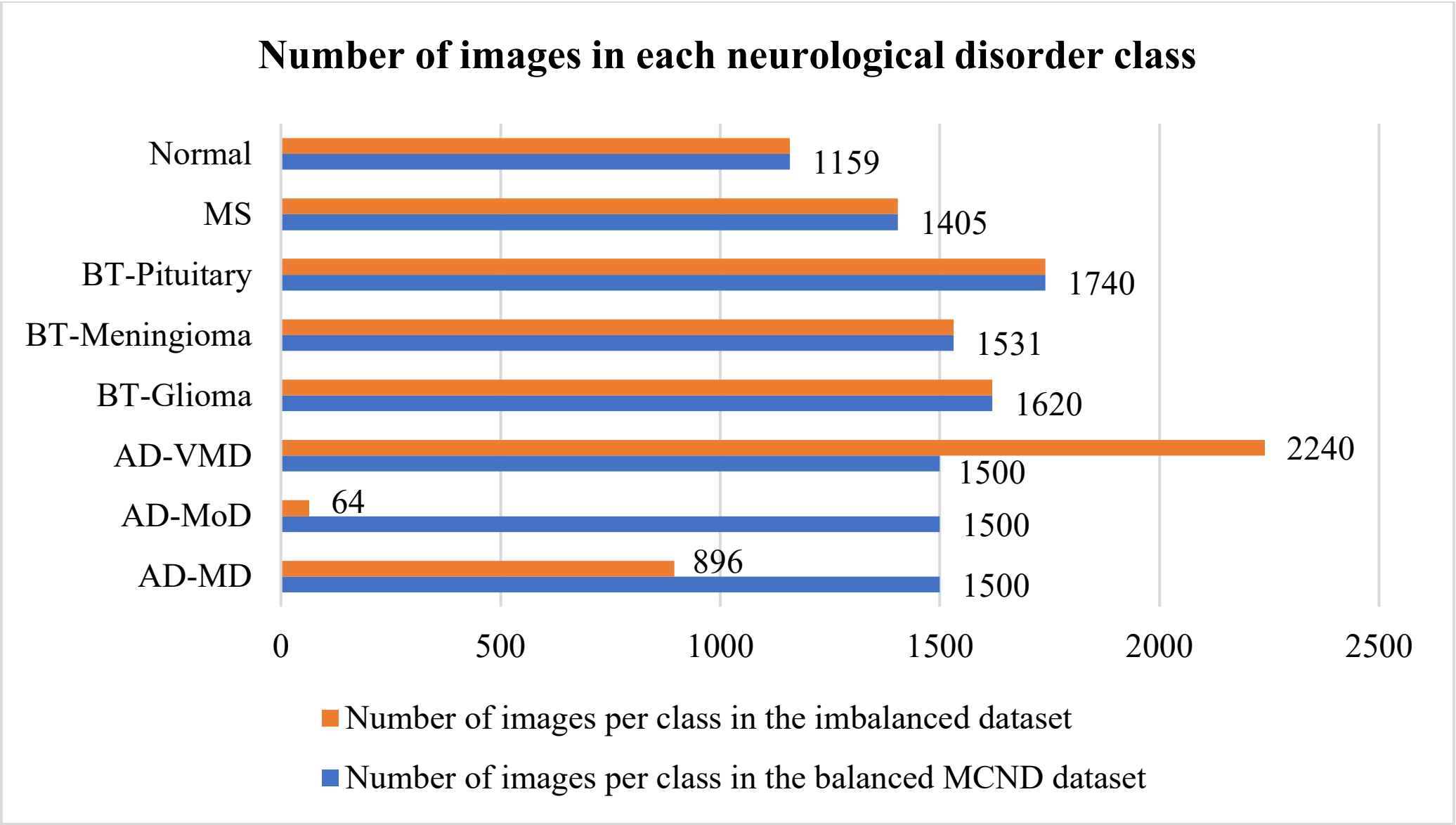

**Fig. 2** Class distribution of the MCND dataset before and after using the WGAN-GP method

**Table 1.** Configuration and hyperparameters of WGAN-GP for oversampling Alzheimer's MRI data

| Hyperparameter / Setting | Value | Purpose / Rationale |
|---|---|---|
| Algorithm | Adam | Faster convergence; helps reduce mode collapse |
| Learning rate | 0.0001 | Stable training for WGAN-GP |
| $\beta_1$ | 0.0 | Common WGAN-GP choice to stabilize updates |
| $\beta_2$ | 0.9 | Momentum on second moment; stable gradients |
| Gradient penalty coefficient (λ) | 10 | Reported to work well across datasets and CNNs |

Table 2 summarizes the composition of the MCND dataset, detailing the number of samples for each class and neurological condition included in the study. The balanced Alzheimer's Disease (AD) dataset contains three classes: AD-MD (1,500 samples), AD-MoD (1,500 samples), and AD-VMD (1,500 samples). The Brain Tumor (BT) dataset includes BT-Glioma (1,620 samples), BT-Meningioma (1,531 samples), and BT-Pituitary (1,740 samples), providing a well-represented distribution across tumor types. The Multiple Sclerosis (MS) dataset comprises 1,405 samples, while 1,159 healthy control MRI scans are sourced from the IXI dataset. This table highlights the diversity and scale of the MCND dataset, which encompasses multiple neurological disorders and healthy controls, providing a robust foundation for training and evaluating deep learning models.

**Table 2.** Details of samples and classes in the MCND dataset

| **Dataset** | **Class Name** | **No. of samples** |
|---|---|---|
| Alzheimer's Disease (AD) dataset [95] | Mild Dementia (AD-MD) | 1,500 |
| | Moderate Dementia (AD-MoD) | 1,500 |
| | Very Mild Dementia (AD-VMD) | 1,500 |
| Brain Tumor (BT) dataset [96] | BT-Glioma | 1,620 |
| | BT-Meningioma | 1,531 |
| | BT-Pituitary | 1,740 |
| Multiple Sclerosis (MS) [97] | MS | 1,405 |
| Information eXtraction from Images (IXI) | Normal (Healthy subjects) | 1,159 |
| Total | 11,955 | |

Fig. 3 illustrates sample data for each class in the MCND dataset, including AD-MD, AD-MoD, AD-VMD, BT-Glioma, BT-Meningioma, BT-Pituitary, MS, and a healthy subject.

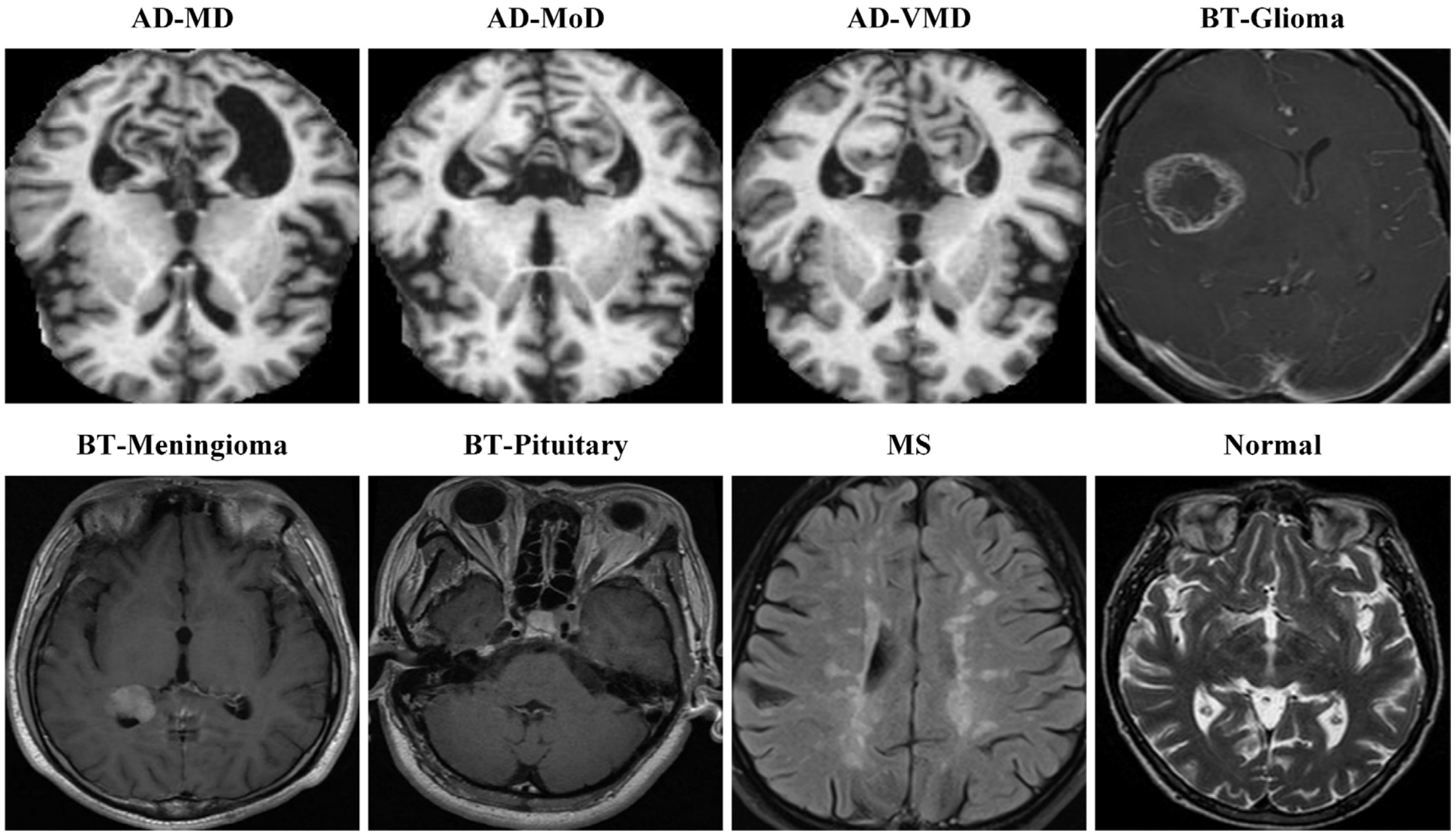

**Fig. 3** Sample data for each class in the MCND dataset

## 5. Experimental Evaluations Phase and Real-Time Deployment

This section presents and compares the proposed End-Net's performance with several well-known deep CNN models, including CNN, AlexNet, VGG19, Inception-V3, DenseNet-121, NeuroNet19, and C-DCNN on the MCND dataset described in Section 4. Table 3 summarizes the findings using performance metrics such as accuracy, recall, F1 score, precision, cross-entropy loss, and MSE. Additionally, the outcomes are visualized in Figs. 4 –15 using accuracy, cross-entropy loss, Receiver Operating Characteristic (ROC), F1 score, MSE curves, Gradient-weighted Class Activation Mapping (Grad-CAM) [103], and confusion matrices. All experiments were conducted in a Kaggle notebook environment, using an NVIDIA Tesla P100 GPU to accelerate processing and ensure efficient performance.

### 5. 1 Performance Metrics

The objective is to benchmark and compare the proposed End-Net's effectiveness across multiple metrics, including accuracy, precision, recall, F1 score, cross-entropy loss, MSE, ROC plot, and confusion matrix. These metrics provide a comprehensive assessment of the model's effectiveness.

The accuracy metric evaluates the ratio of accurate forecasts generated by the model and is computed using Eq. (13) [104], where TP, TN, FP, and FN denote the counts of true positives, true negatives, false positives, and false negatives, respectively.

$$Accuracy = \frac{TP + TN}{TP + FN + FP + TN} \tag{13}$$

The precision metric is the ratio of true positives to the total number of positive predictions. It is calculated based on Eq. (14) [104].

$$Precision = \frac{TP}{TP + FP} \tag{14}$$

The recall metric measures the proportion of relevant instances a model correctly identifies out of the total number of relevant instances in the dataset. It is calculated using Eq. (15) [104].

$$\text{Recall} = \frac{TP}{TP + FN} \tag{15}$$

The F1 score combines precision and recall into a unified measure and is calculated using Eq. (16) [104].

$$F1\ score = 2 \times \frac{Precision \times Recall}{Precision + Recall} \tag{16}$$

The cross-entropy loss metric evaluates the mathematical discrepancy between predicted and actual values. It is computed using Eq. (17) [105], where *n* denotes the total number of classes, and yi is a binary indicator indicating whether class *i* correctly classifies the observation. $P_i$ denotes the predicted probability for class *i*.

$$Cross - entorpy\ loss = -\sum_{i=1}^{n} y_i \log(p_i) \tag{17}$$

The MSE serves as another loss function that measures the squared deviation between predicted $\hat{y}_i$ and actual values $y_i$ using Eq. (18).

$$MSE = \frac{1}{N}\sum_{i=1}^{n}(y_i - \hat{y}_i)^2 \tag{18}$$

The ROC curve plots the true positive rate (recall) on the y-axis against the false positive rate (1 - specificity) on the x-axis as classification thresholds vary. These rates are calculated using Eq. (19) [106].

$$FRP = \frac{FP}{FP + TN}$$
$$TRP = \frac{TP}{TP + FN} \tag{19}$$

The confusion matrix summarizes prediction outcomes, categorizing correct and incorrect predictions for each class.

## 5. 2 Experimental Settings and Hyper-parameter Configurations

To ensure consistency across datasets and compatibility, the input dimensions for the proposed End-Net and the CNN model were set to 210 × 210 pixels. For other models, the input image sizes were as follows: AlexNet (227×227), VGG19 (224×224), Inception-V3 (299×299), DenseNet-121 (224×224), NeuroNet19 (128×128), and C-DCNN (64×64), as specified in their respective papers. Additionally, all images were normalized to standardize pixel values, and categorical labels were converted to one-hot encoding for numerical suitability. The models' performance was assessed using 5-fold cross-validation, with the dataset partitioned into 5 equal subsets for robust evaluation. In this process, each fold served in turn as the test set, with half of the data further divided to create a validation set, and the remaining folds allocated for training. For the VGG19, Inception-V3, DenseNet-121, and NeuroNet19 models, transfer learning was applied by fine-tuning the last three layers. This approach adapted the pre-trained models to the specific dataset, enabling them to retain learned features while optimizing performance for the task at hand.

Various hyperparameters were carefully calibrated through controlled experiments to enhance the model's predictive capabilities and its ability to effectively classify neurological disorders. These parameters included batch size, optimizer, learning rate, number of epochs, and the loss function. Since the MCND dataset required classification across eight distinct classes, categorical cross-entropy was used as the loss function because it is mathematically well-suited to multi-class classification tasks. The Adam optimizer [107] is utilized for the proposed End-Net, and all seven contender models are initialized with a learning rate of 0.001 ($1e^{-3}$). To further optimize the training process, the learning rate was decayed by 0.1 after every 5 epochs, contingent on the validation loss, with a lower bound of 0.0000001 ($1e^{-7}$). The optimized hyperparameter values utilized throughout the experiments are tabulated in Table 3. These configurations were finalized following extensive fine-tuning and a series of experimental evaluations.

**Table 3.** Summary of optimized hyperparameter settings for models.

| **Parameters** | **Optimized value** |
|---|---|
| Input shape | (210, 210, 3) |
| Optimizer | Adam |
| Initial learning rate | 0.001 |
| Patience | 5 |
| Learning rate decay factor | 0.1 |
| Learning rate lower bound | $1e^{-7}$ |
| Hidden layer activation | ReLU |
| Number of epochs | 40 |
| Batch size | 32 |
| Dropout rate | 0.3 |
| Dense layers | 512 → 256 → 8 |
| Output layer activation function | Softmax |

Furthermore, a consistent architecture for the dense layers was implemented across all eight models to ensure comparability and optimize performance. As mentioned in Section 3.3, the dense layer configuration included global average pooling to aggregate spatial features, followed by a 512-unit dense layer with ReLU activation and a dropout layer with a dropout rate of 0.3. A 256-unit dense layer with ReLU and another 0.3 dropout layer were added for feature abstraction and regularization. The last dense layer comprised eight units with softmax activation, outputting class probabilities that balanced complexity and generalization for multi-class classification.

### 5. 3 End-Net Ablation Study

To evaluate the contribution of the proposed End-Net to domain generalization, we conducted an ablation study using three architectures: a version of End-Net without the Inception module End-Net (No Inception Module), Inception-V3, and the full proposed End-Net. The End-Net–No Inception serves as a simplified reference, while Inception-V3 captures deeper, multiscale features through its inception modules. The full End-Net integrates tailored architectural enhancements to improve robustness across datasets for neurological disorders.

The validation accuracy comparison in the ablation study is presented in Fig. 4. Validation accuracy across 40 epochs was evaluated for three models: End-Net without the Inception module (End-Net–No Inception), Inception-V3, and the full proposed End-Net. End-Net–No Inception started with low accuracy (~0.16) and gradually improved to ~0.93, indicating slower convergence and limited multiscale feature learning. Inception-V3 began around 0.81 and steadily increased to 0.91, benefiting from multiscale feature extraction, but achieved a slightly lower final accuracy than End-Net. The full End-Net surpassed both models quickly, reaching ~0.95 by epoch 15 and plateauing near 0.972, highlighting the effectiveness of its architectural enhancements in improving learning speed, final accuracy, and robustness across neurological disorder datasets.

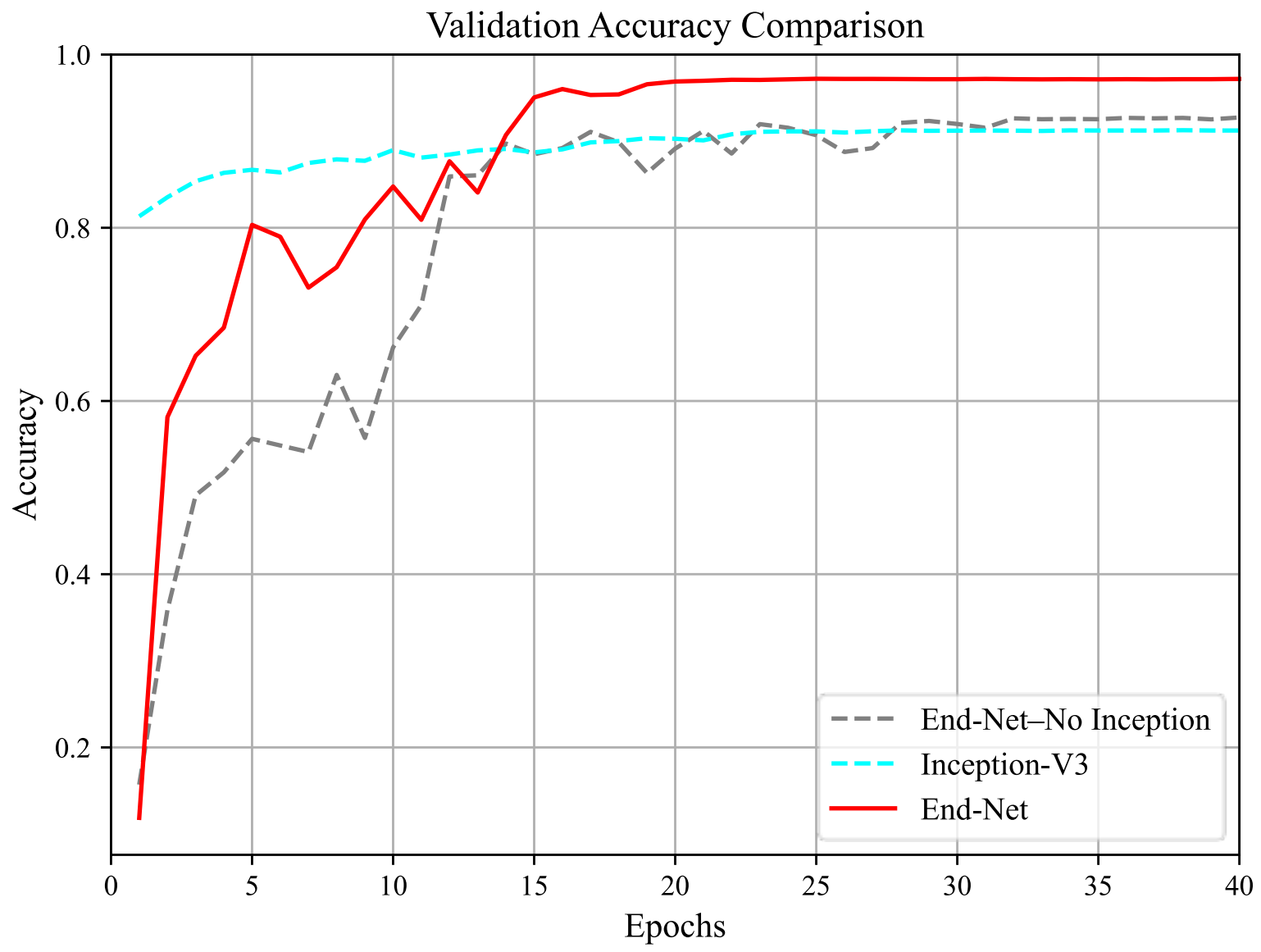

**Fig. 4** Comparison of validation accuracy in the ablation study

The validation F1-score comparison in the ablation study is shown in Fig. 5. End-Net–No Inception started with a very low F1 score (~0.15) and gradually increased to ~0.927, indicating slower convergence and limited multiscale feature learning. Inception-V3 began around 0.81 and steadily improved to 0.912, benefiting from multiscale feature extraction, but achieved slightly lower final performance than End-Net. The full End-Net demonstrated rapid gains, surpassing the other models by the middle epochs (~0.95 by epoch 15) and plateauing near 0.972, highlighting that its architectural enhancements significantly improve the balance between precision and recall, thereby enhancing overall robustness across neurological disorder datasets.

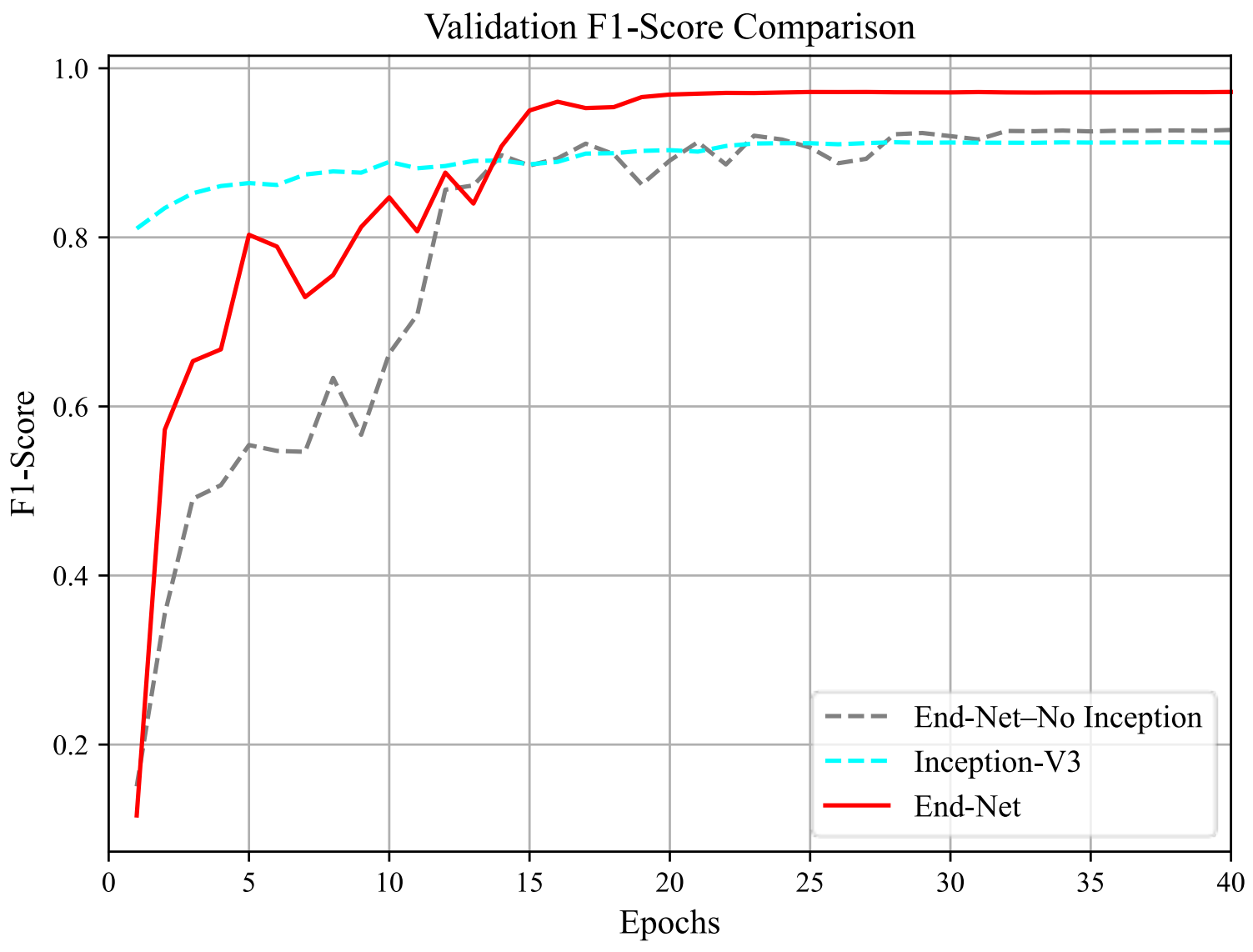

**Fig. 5** Validation F1 score comparison in the ablation study

The validation loss comparison in the ablation study is shown in Fig. 6. End-Net–No Inception started with a very high loss (~7.91) and gradually decreased to around 0.18, indicating slow convergence and limited feature-learning capacity. Inception-V3 began with a lower initial loss (~0.43) and steadily decreased to approximately 0.267, showing better convergence due to its multiscale feature extraction, but still had a slightly higher final loss than End-Net. The full End-Net exhibited rapid loss reduction, starting at ~7.55 and quickly reaching ~0.14 by the middle epochs, then plateauing near this value, highlighting the effectiveness of its architectural enhancements in accelerating convergence and improving overall learning stability across neurological disorder datasets.

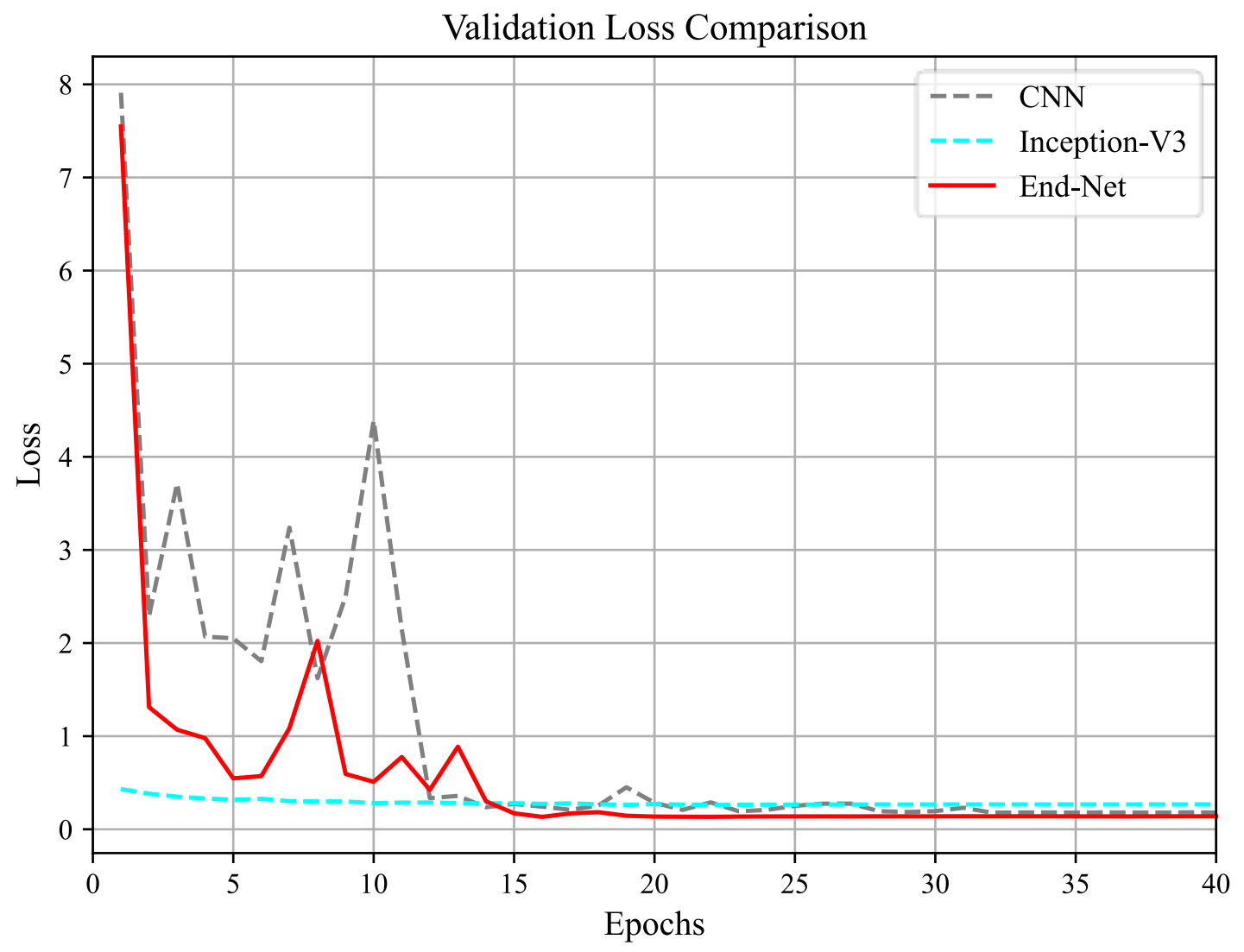

**Fig. 6** Validation loss comparison in the ablation study

## 5. 4 Results of the Proposed End-Net

This section evaluates the proposed End-Net's generalization capacity by assessing the trade-off between overfitting and underfitting. The performance metrics in Fig. 7 indicate that the proposed End-Net attains high accuracy on the training dataset. Furthermore, it performs well on the validation set, suggesting it has effectively learned the underlying patterns without overfitting the MCND training data. The observed balance in accuracy between the training and validation sets suggests the model's robust generalization to unseen instances, underscoring its potential reliability for classifying neurological disorders in the studied dataset.

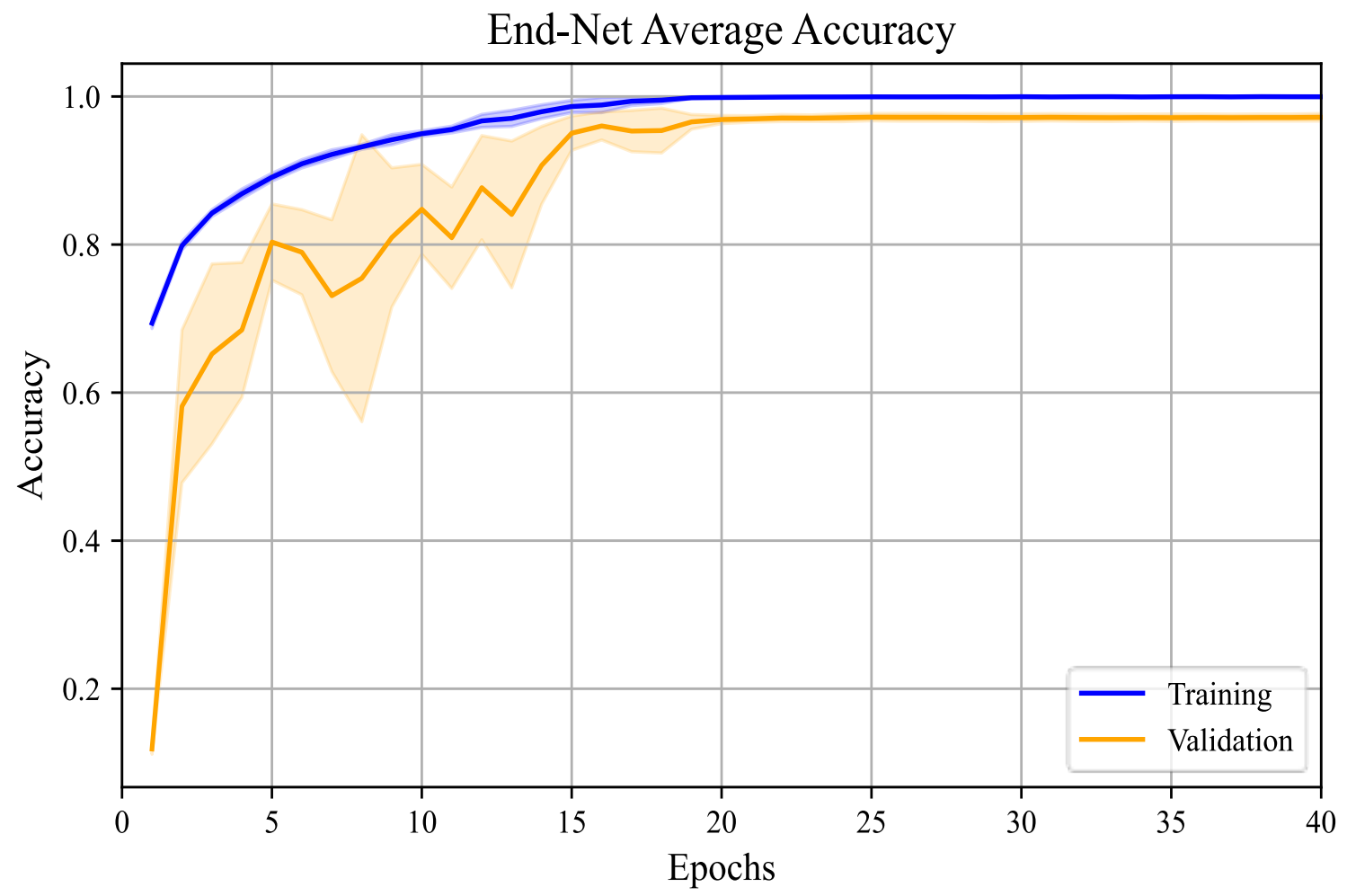

**Fig. 7** Comparison of training and validation accuracy results of the End-Net on the MCND dataset

The results in Fig. 8 analyze the cross-entropy loss, providing essential insights into the model's performance. A consistent decrease in the training and validation loss curves indicates effective learning and successful management of overfitting. The results indicate that the model effectively captures the underlying patterns and relationships within the training dataset and iteratively adjusts its parameters to optimize predictions on unseen data, further validating its robustness. The consistently low validation cross-entropy loss underscores the model's ability to generalize effectively in detecting neurological disorders.

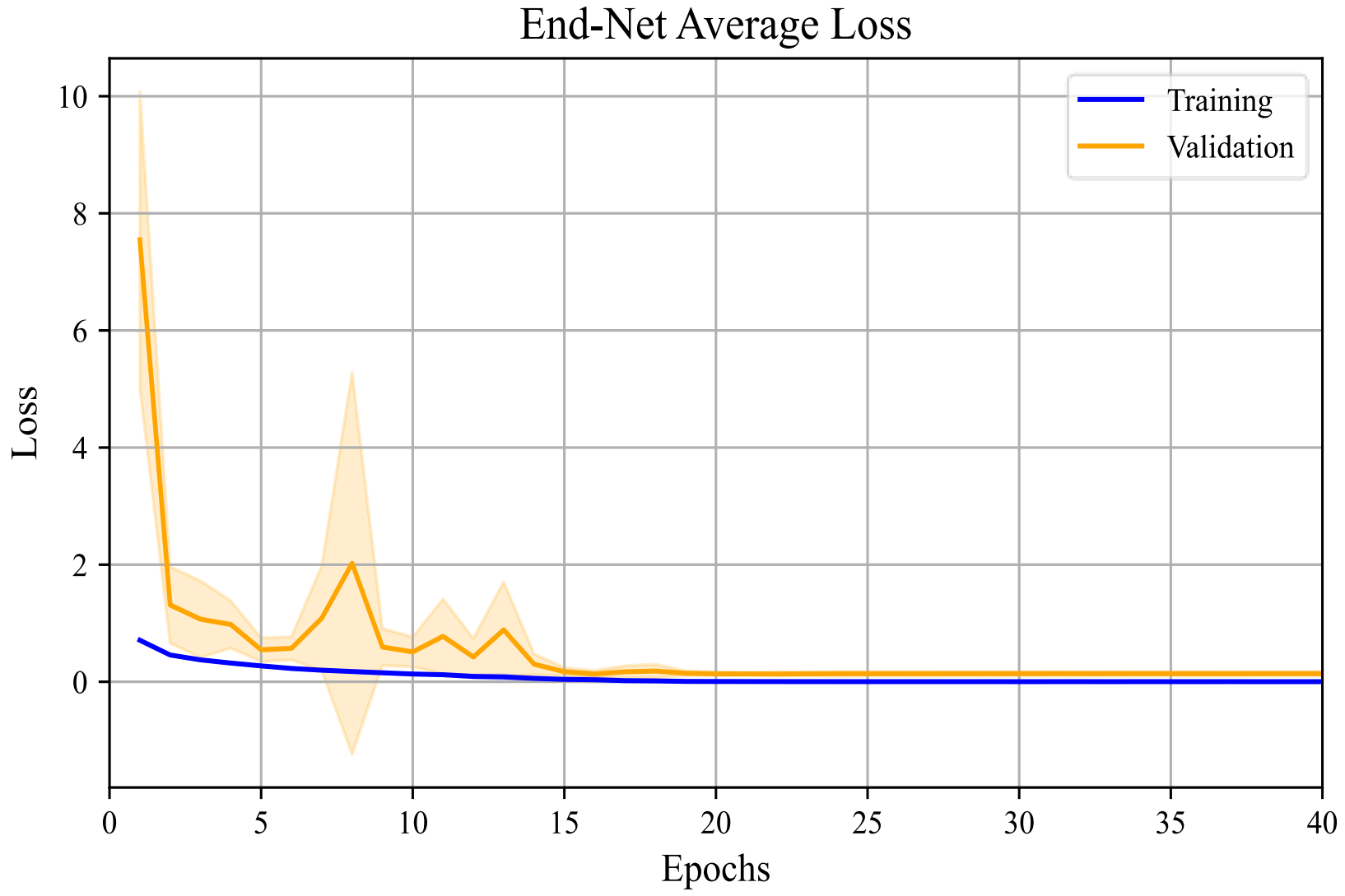

**Fig. 8** Comparison of cross-entropy loss results of the End-Net on the MCND dataset

ROC curves are a fundamental evaluation tool in binary classification, graphically illustrating the balance between a classifier's sensitivity and specificity. The output must be binarized to broaden the ROC curve and ROC area for multi-class classification. Fig. 9 depicts the extended ROC curve of the proposed End-Net, highlighting its strong predictive performance across multiple classes. The curves demonstrate a high True Positive Rate (TPR) relative to the False Positive Rate (FPR) across various thresholds. The obtained results indicate that the proposed model effectively balances recall and specificity.

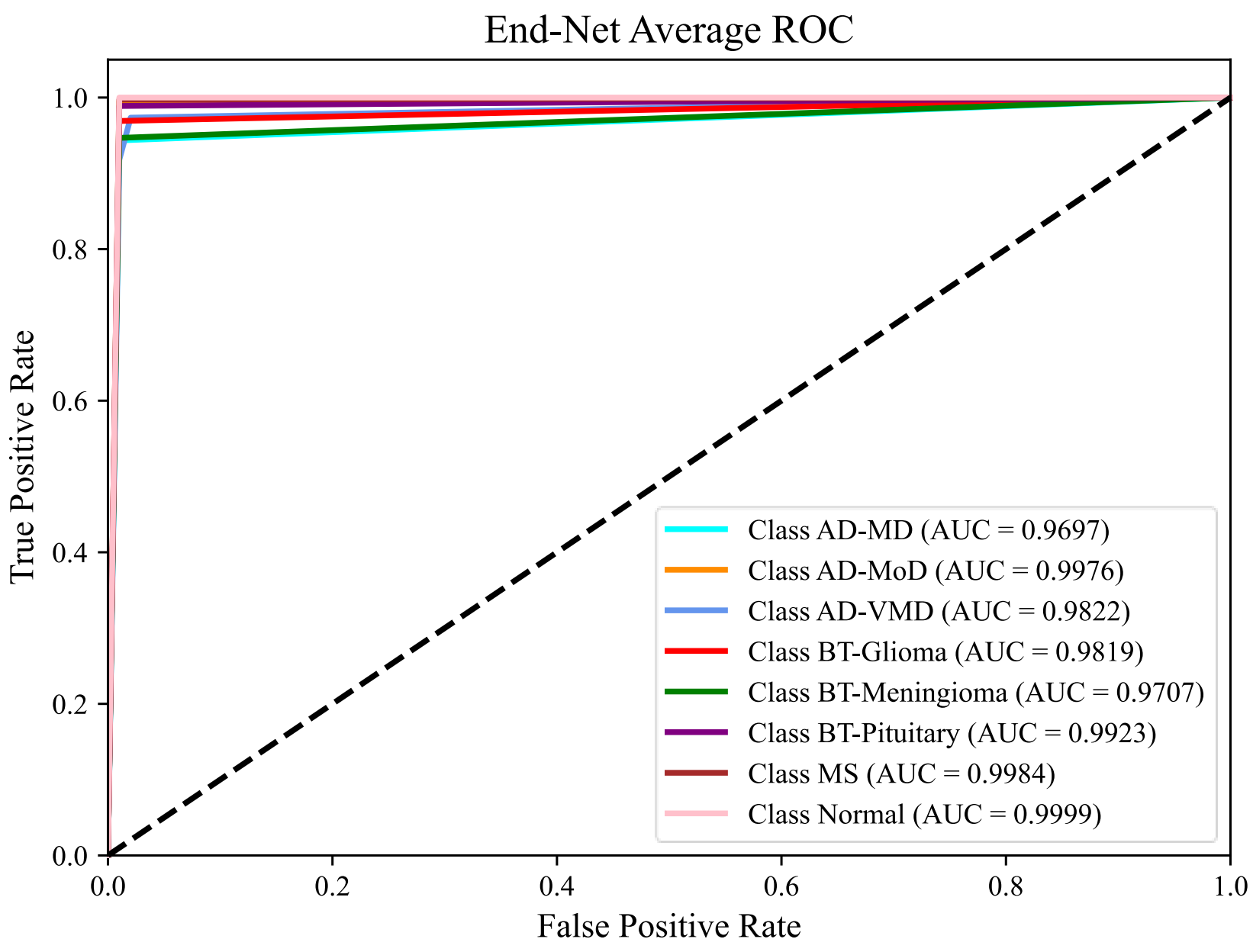

**Fig. 9** Average ROC curve of the proposed End-Net on the MCND dataset

## 5. 5 Performance Comparison of End-Net with State-of-the-art Deep Models

In this subsection, two sets of experiments were conducted to assess and compare the proposed End-Net model with seven established deep learning models: CNN, AlexNet [20], VGG19 [21], Inception-V3 [33], DenseNet-121 [25], NeuroNet19 [26], and C-DCNN [28]. In the first experiment, the performance of the proposed End-Net model, along with seven benchmark deep learning models, was systematically evaluated on separate datasets for neurological disorders, including Alzheimer's disease, brain tumors, and multiple sclerosis. This evaluation aimed to examine how effectively each model learned and made predictions when trained and tested within a specific domain of disorders. Conducting experiments on these distinct datasets provided insight into the models' capacity to capture disorder-specific patterns and features, as well as their robustness to variations within each condition. The detailed results of this experiment, including comparative performance metrics for all models, are summarized in Tables 4–6. In the second experiment, the models' generalization ability was evaluated using the MCND dataset, which includes multiple neurological disorders. The results of this evaluation are reported in Table 7. For all tables, comparisons were made using key evaluation metrics: accuracy, precision, recall, F1 score, cross-entropy loss, and MSE, with the best-performing values highlighted in bold.

Table 4 compares the proposed End-Net model against several leading deep learning models for Alzheimer's disease classification. Across all evaluation metrics—accuracy, precision, recall,

F1 score, cross-entropy loss, and MSE—End-Net consistently outperforms the other models. Specifically, End-Net achieves the highest accuracy, precision, recall, and F1 score, all at 0.9850, reflecting superior classification performance and balanced class-balanced diagnosis. In addition, it achieves the lowest cross-entropy loss (0.0625) and MSE (0.0068), demonstrating improved diagnostic reliability and lower error than CNN, AlexNet, VGG19, InceptionV3, DenseNet, NeuroNet19, and C-DCNN. These results underscore the End-Net model's ability to effectively capture distinctive features of Alzheimer's disease, enabling more accurate and robust diagnoses.

**Table 4.** Comparison of the proposed End-Net performance against leading models in AD.

| Models | Statistical Metrics | | | | | |
|---|---|---|---|---|---|---|
| | **Accuracy** | **Precision** | **Recall** | **F1 score** | **Cross-entropy loss** | **MSE** |
| CNN | 0.9583 | 0.9583 | 0.9583 | 0.9583 | 0.1008 | 0.0163 |
| AlexNet | 0.8800 | 0.8959 | 0.8750 | 0.8853 | 0.2592 | 0.0404 |
| VGG19 | 0.9783 | 0.9800 | 0.9783 | 0.9791 | 0.1012 | 0.0096 |
| InceptionV3 | 0.9133 | 0.9133 | 0.9133 | 0.9133 | 0.2244 | 0.0317 |
| DenseNet | 0.9700 | 0.9700 | 0.9700 | 0.9700 | 0.0889 | 0.0126 |
| NeuroNet19 | 0.9467 | 0.9466 | 0.9450 | 0.9458 | 0.1970 | 0.0210 |
| C-DCNN | 0.9633 | 0.9633 | 0.9617 | 0.9625 | 0.0923 | 0.0146 |
| End-Net | **0.9850** | **0.9850** | **0.9850** | **0.9850** | **0.0625** | **0.0068** |

Table 5 provides a comparative analysis of the proposed End-Net model against several state-of-the-art deep learning models for brain tumor (BT) classification. End-Net achieves the highest accuracy (0.9743), precision (0.9758), recall (0.9743), and F1 score (0.9750), demonstrating excellent classification performance and balanced diagnosis across all classes. Furthermore, it attains the lowest cross-entropy loss (0.0835) and MSE (0.0095), indicating high diagnostic reliability and minimal error compared to CNN, AlexNet, VGG19, InceptionV3, DenseNet, NeuroNet19, and C-DCNN.

**Table 5.** Comparison of the proposed End-Net performance against leading models in BT.

| Models | Statistical Metrics | | | | | |
|---|---|---|---|---|---|---|
| | Accuracy | Precision | Recall | F1 score | Cross-entropy loss | MSE |
| CNN | 0.9395 | 0.9408 | 0.9380 | 0.9394 | 0.1591 | 0.0227 |
| AlexNet | 0.8880 | 0.8894 | 0.8880 | 0.8887 | 0.4164 | 0.0454 |
| VGG19 | 0.9561 | 0.9561 | 0.9561 | 0.9561 | 0.2222 | 0.0203 |
| InceptionV3 | 0.9198 | 0.9212 | 0.9198 | 0.9205 | 0.3008 | 0.0327 |
| DenseNet | 0.9486 | 0.9514 | 0.9471 | 0.9492 | 0.1854 | 0.0206 |
| NeuroNet19 | 0.9228 | 0.9255 | 0.9213 | 0.9234 | 0.3869 | 0.0305 |
| C-DCNN | 0.9410 | 0.9452 | 0.9395 | 0.9423 | 0.2965 | 0.0241 |
| End-Net | **0.9743** | **0.9758** | **0.9743** | **0.9750** | **0.0835** | **0.0095** |

Table 6 compares the proposed End-Net model against several leading deep learning models for MS classification. End-Net achieves the highest accuracy (0.9360), precision (0.9360), recall (0.9360), and F1 score (0.9360), demonstrating superior overall classification performance. While its cross-entropy loss (0.2611) is slightly higher than that of CNN (0.2400), End-Net achieves the lowest mean squared error (0.0522), indicating reliable predictions with minimal error. Compared to AlexNet, VGG19, InceptionV3, DenseNet, NeuroNet19, and C-DCNN, End-Net provides the most balanced and robust MS diagnosis across metrics.

**Table 6.** Comparison of the proposed End-Net performance against leading models in MS.

| Models | Statistical Metrics | | | | | |
|---|---|---|---|---|---|---|
| | Accuracy | Precision | Recall | F1 score | Cross-entropy loss | MSE |
| CNN | 0.8953 | 0.8953 | 0.8953 | 0.8953 | **0.2400** | 0.0714 |
| AlexNet | 0.8256 | 0.8256 | 0.8256 | 0.8256 | 0.4927 | 0.1352 |
| VGG19 | 0.9070 | 0.9070 | 0.9070 | 0.9070 | 0.4048 | 0.0796 |
| InceptionV3 | 0.8488 | 0.8488 | 0.8488 | 0.8488 | 0.3869 | 0.1185 |
| DenseNet | 0.8837 | 0.8837 | 0.8837 | 0.8837 | 0.2702 | 0.0813 |
| NeuroNet19 | 0.8953 | 0.8953 | 0.8953 | 0.8953 | 0.3132 | 0.0845 |
| C-DCNN | 0.9215 | 0.9215 | 0.9215 | 0.9215 | 0.4019 | 0.0647 |
| End-Net | **0.9360** | **0.9360** | **0.9360** | **0.9360** | 0.2611 | **0.0522** |

Table 7 presents the results of the second experiment conducted on the MCND dataset, which encompasses multiple neurological disorders. The proposed End-Net model achieved the highest overall performance across all evaluation metrics, demonstrating robust performance across diverse neurological conditions. Specifically, End-Net achieved an accuracy of 0.9761, outperforming VGG19 (0.9676) and AlexNet (0.9547), indicating its superior ability to correctly classify cases across different disorders. In terms of precision and recall, End-Net achieved 0.9766 and 0.9761, respectively, demonstrating a strong ability to correctly identify positive cases while minimizing false positives and false negatives. Its F1 score of 0.9763 further emphasizes the model's balanced performance, effectively integrating precision and recall into a single metric. Regarding optimization performance, End-Net achieved the lowest cross-entropy loss (0.1024) compared to DenseNet (0.1698) and VGG19 (0.1929), indicating faster convergence and improved model learning. Similarly, End-Net achieved the smallest mean squared error (0.0050), outperforming VGG19 (0.0071) and DenseNet (0.0107), demonstrating superior diagnostic accuracy and stability when applied to a dataset comprising multiple neurological disorders. Overall, these results highlight End-Net's capability to provide reliable and precise diagnoses across a range of complex neurological conditions.

**Table 7.** Comparison of the proposed End-Net performance against leading models on the MCND dataset.

| **Metrics** | | **CNN** | **AlexNet** | **VGG19** | **InceptionV3** | **DenseNet** | **NeuroNet19** | **C-DCNN** | **End-Net** |
|---|---|---|---|---|---|---|---|---|---|
| Accuracy | Mean | 0.9343 | 0.9547 | 0.9676 | 0.9189 | 0.9445 | 0.9221 | 0.8567 | **0.9761** |
| | Std | 0.0106 | 0.0049 | 0.0057 | 0.0089 | 0.0062 | 0.0059 | 0.2060 | 0.004 |
| Precision | Mean | 0.9359 | 0.9550 | 0.968 | 0.9195 | 0.9456 | 0.924 | 0.9475 | **0.9766** |
| | Std | 0.0099 | 0.0044 | 0.0051 | 0.0094 | 0.0062 | 0.0065 | 0.0074 | 0.0043 |
| Recall | Mean | 0.9331 | 0.9547 | 0.9671 | 0.9171 | 0.944 | 0.9204 | 0.8286 | **0.9761** |
| | Std | 0.0112 | 0.0049 | 0.0052 | 0.0088 | 0.0064 | 0.0069 | 0.2660 | 0.004 |
| Specificity | Mean | 0.9906 | 0.9935 | 0.9953 | 0.9884 | 0.9921 | 0.9889 | 0.9794 | **0.9966** |
| | Std | 0.0015 | 0.0007 | 0.0008 | 0.0012 | 0.0009 | 0.0008 | 0.0298 | 0.0006 |
| F1 score | Mean | 0.9345 | 0.9548 | 0.9675 | 0.9183 | 0.9448 | 0.9222 | 0.8617 | **0.9763** |
| | Std | 0.0105 | 0.0046 | 0.0051 | 0.0091 | 0.0063 | 0.0066 | 0.1953 | 0.0042 |
| Cross-entropy loss | Mean | 0.154 | 0.2309 | 0.1929 | 0.2343 | 0.1698 | 0.2541 | 0.3526 | **0.1024** |
| | Std | 0.028 | 0.0466 | 0.0552 | 0.0212 | 0.0122 | 0.0345 | 0.3890 | 0.0203 |
| MSE | Mean | 0.0114 | 0.0097 | 0.0071 | 0.0151 | 0.0107 | 0.0149 | 0.0213 | **0.005** |
| | Std | 0.0018 | 0.0011 | 0.0013 | 0.0017 | 0.0012 | 0.0015 | 0.0250 | 0.0009 |
| Friedman's average rank | | 5.6 | 3.2 | 2 | 7.6 | 4.6 | 7 | 5 | **1** |

Fig. 10 illustrates the F1 score performance of End-Net compared to other deep learning models on the MCND dataset. End-Net achieves the highest overall F1 score of approximately 0.9720, outperforming VGG19 (0.9616), AlexNet (0.9457), Inception-V3 (0.9125), DenseNet-121 (0.9391), NeuroNet19 (0.9130), and C-DCNN (0.8618). Starting from an initial F1 score of 0.116, End-Net rapidly exceeds 0.90 and reaches 0.9720, demonstrating a steeper and more sustained learning curve compared to other models, which either improve more gradually or plateau at lower values. Although VGG19 and DenseNet-121 show competitive performance, their curves flatten earlier, and their final F1 scores remain below End-Net's. These results highlight End-Net's superior ability to achieve high, stable diagnostic performance across multiple neurological disorders in the MCND dataset.

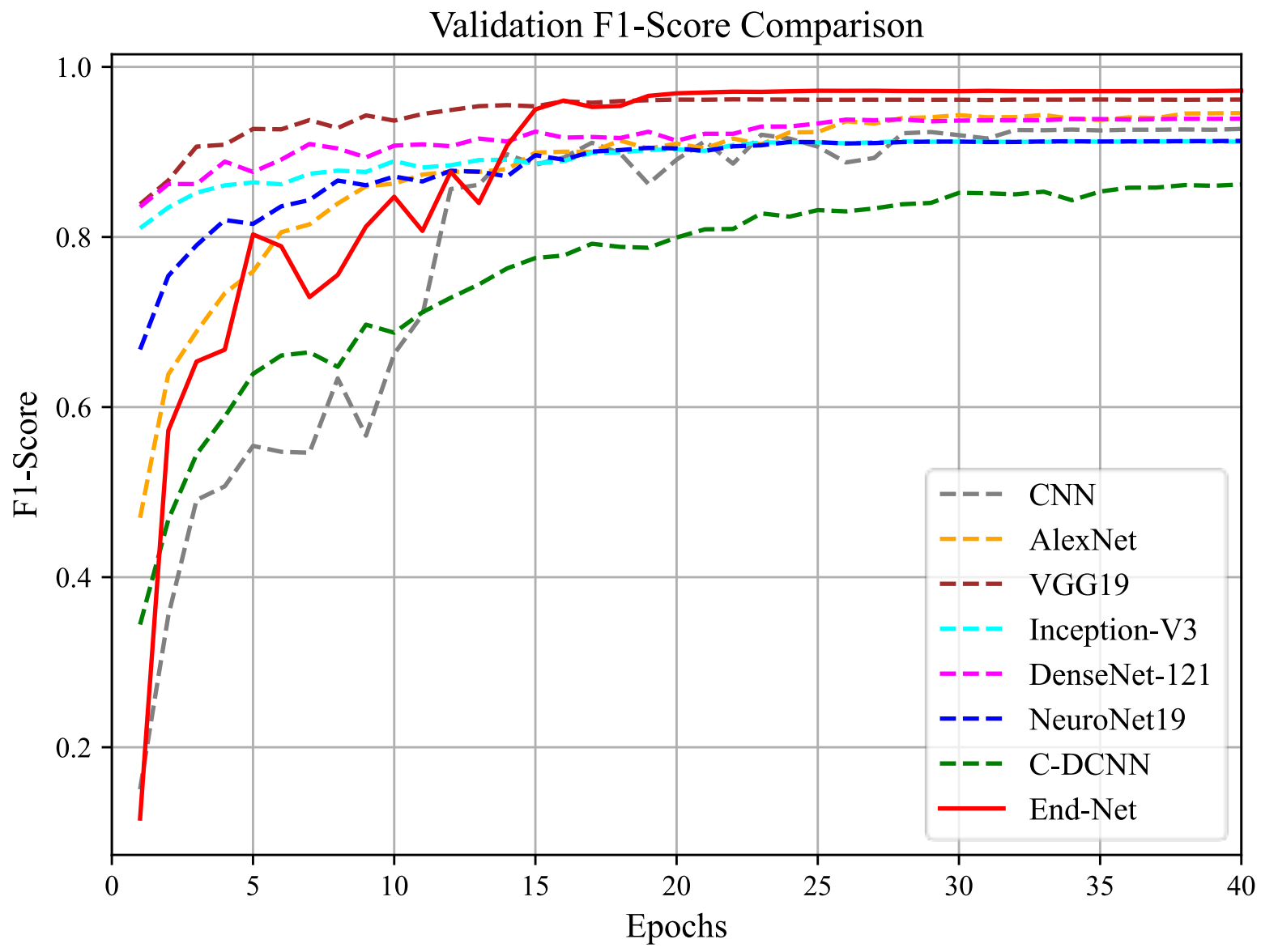

**Fig. 10** Comparison of F1 scores between the proposed End-Net and the competing models on the MCND dataset.

Fig. 11 illustrates the validation results for the proposed End-Net and competing models on the MCND dataset. End-Net shows the most pronounced improvement, starting with an initial MSE of 0.2022 and reducing it to approximately 0.0061 by the final epochs, indicating superior learning and minimal residual error. In comparison, CNN begins at 0.1859 and converges to approximately 0.0129, demonstrating substantial improvement but still less effective than End-Net. AlexNet starts at 0.0634 and declines steadily to 0.0117, while VGG19 begins at 0.0280 and reaches approximately 0.0084, demonstrating a significant reduction in error. Inception-V3 decreases from

0.0317 to 0.0169, DenseNet-121 from 0.0275 to 0.0114, NeuroNet19 from 0.0485 to 0.0169, and C-DCNN from 0.0716 to 0.0219, all showing moderate improvements.

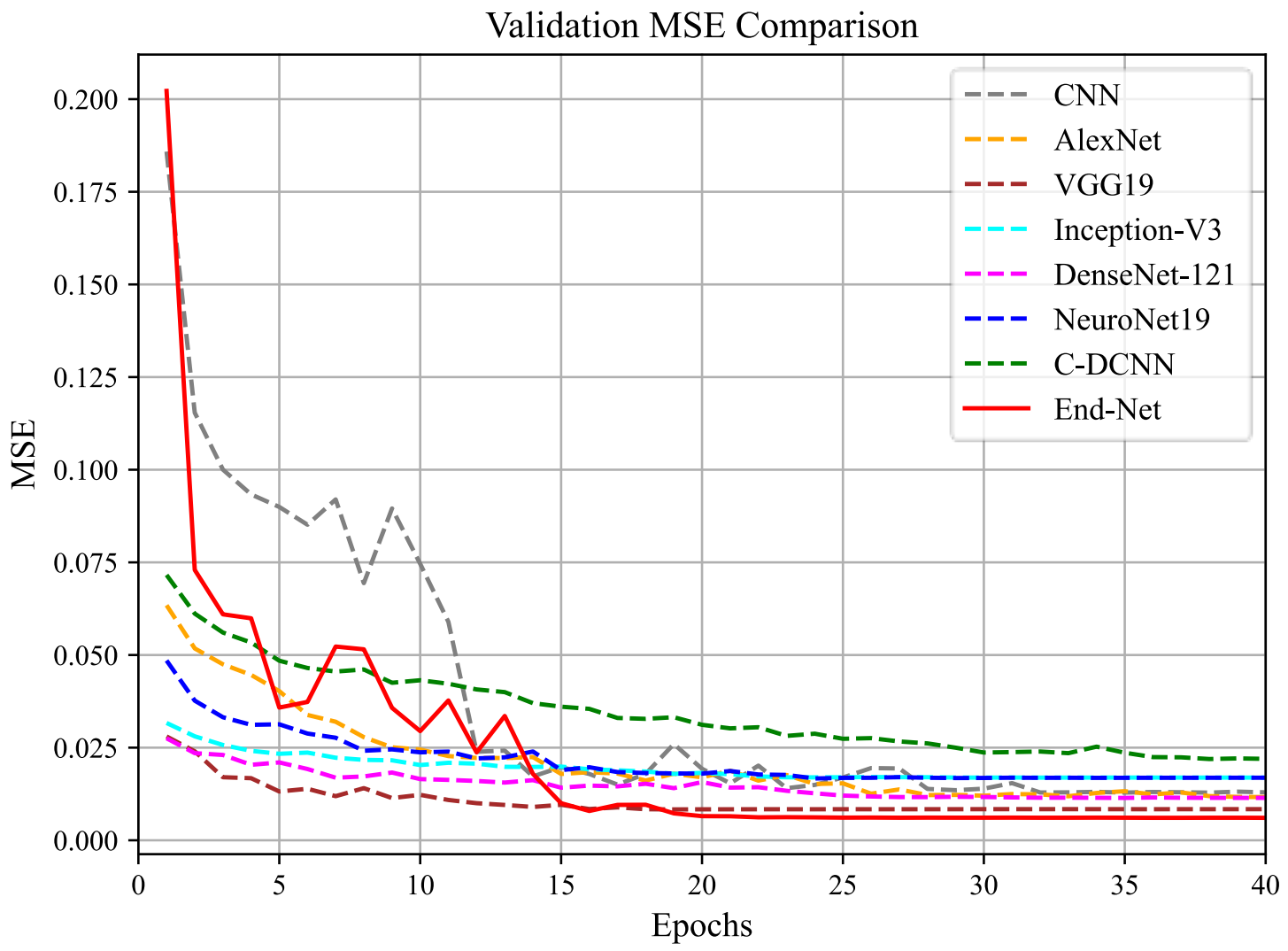

**Fig. 11** MSE comparison of the proposed End-Net and the competing models on the MCND dataset.

Fig. 12 compares the ROC curves of End-Net with other deep learning models on the MCDN dataset. Across the range of decision thresholds, End-Net consistently demonstrates higher True Positive Rates (TPRs), indicating superior recall. End-Net's TPR starts at approximately 0.976 and remains the highest across all thresholds, highlighting its robust detection capability. VGG19 closely follows, starting near 0.968, while AlexNet begins around 0.955. CNN and DenseNet-121 show moderate initial performance, starting at 0.914 and 0.945, respectively, and steadily improve. NeuroNet19 and Inception-V3 start lower, at approximately 0.841 and 0.804, respectively, while C-DCNN begins near 0.773. Although all models eventually reach a TPR of 1.0 at high thresholds, End-Net consistently outperforms the others, demonstrating its superior ability to capture positive instances and maintain high recall even in early classification stages.

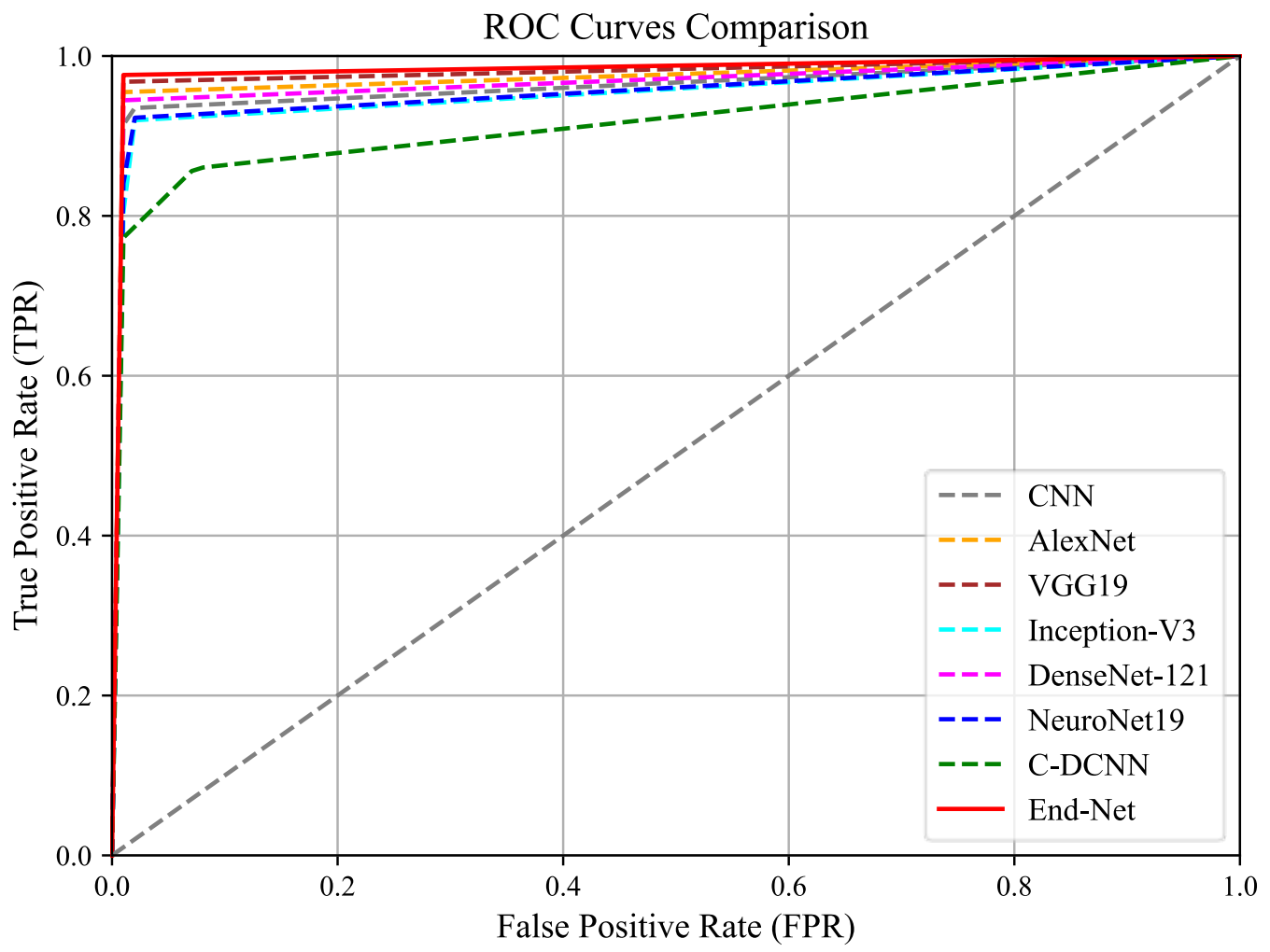

**Fig. 12** ROC comparison of the proposed End-Net and the competing models on the MCND dataset

The confusion matrices in Fig. 13 provide the raw counts and row-wise normalized classification results for the proposed End-Net and seven competing models. Each matrix is normalized by dividing each cell's value by the total number of samples in its respective row, converting raw counts into proportions between 0% and 100%. End-Net demonstrates superior class-specific recall across most categories, achieving 94.30% for AD-MD (correctly classifying 94.30% of true AD-MD cases) and 97.31% for AD-VMD, outperforming all other models. By contrast, Inception-V3 and DenseNet-121 achieve recalls of only 80.26% and 91.04% for AD-MD, respectively, reflecting higher misclassification rates. End-Net also excels in brain tumor categories, with 96.89% recall for BT-Glioma, 94.65% for BT-Meningioma, and 98.88% for BT-Pituitary. In comparison, models such as AlexNet and NeuroNet19 show lower recall for BT-Meningioma (91.21% and 86.24%, respectively) than End-Net. These results demonstrate End-Net's robustness in distinguishing among multiple neurological disorders despite the challenges posed by dataset imbalance.

However, the proposed End-Net shows lower dominance in AD-MoD classification, achieving a recall of 99.52%, matching VGG19 (99.52%) and slightly exceeding C-DCNN (99.04%). For MS classification, End-Net achieves a recall of 99.73%, surpassing CNN (99.59%), DenseNet-121 (99.32%), Inception-V3 (98.36%), and C-DCNN (79.32%). This highlights End-Net's superior ability to capture disease-specific features compared to other architectures. Notably, the End-Net, CNN, and AlexNet achieve the highest recall for Normal cases (100%), effectively minimizing false positives, while Inception-V3 (99.83%), VGG19 (99.67%), and C-DCNN (80.13%) achieve slightly lower recall.

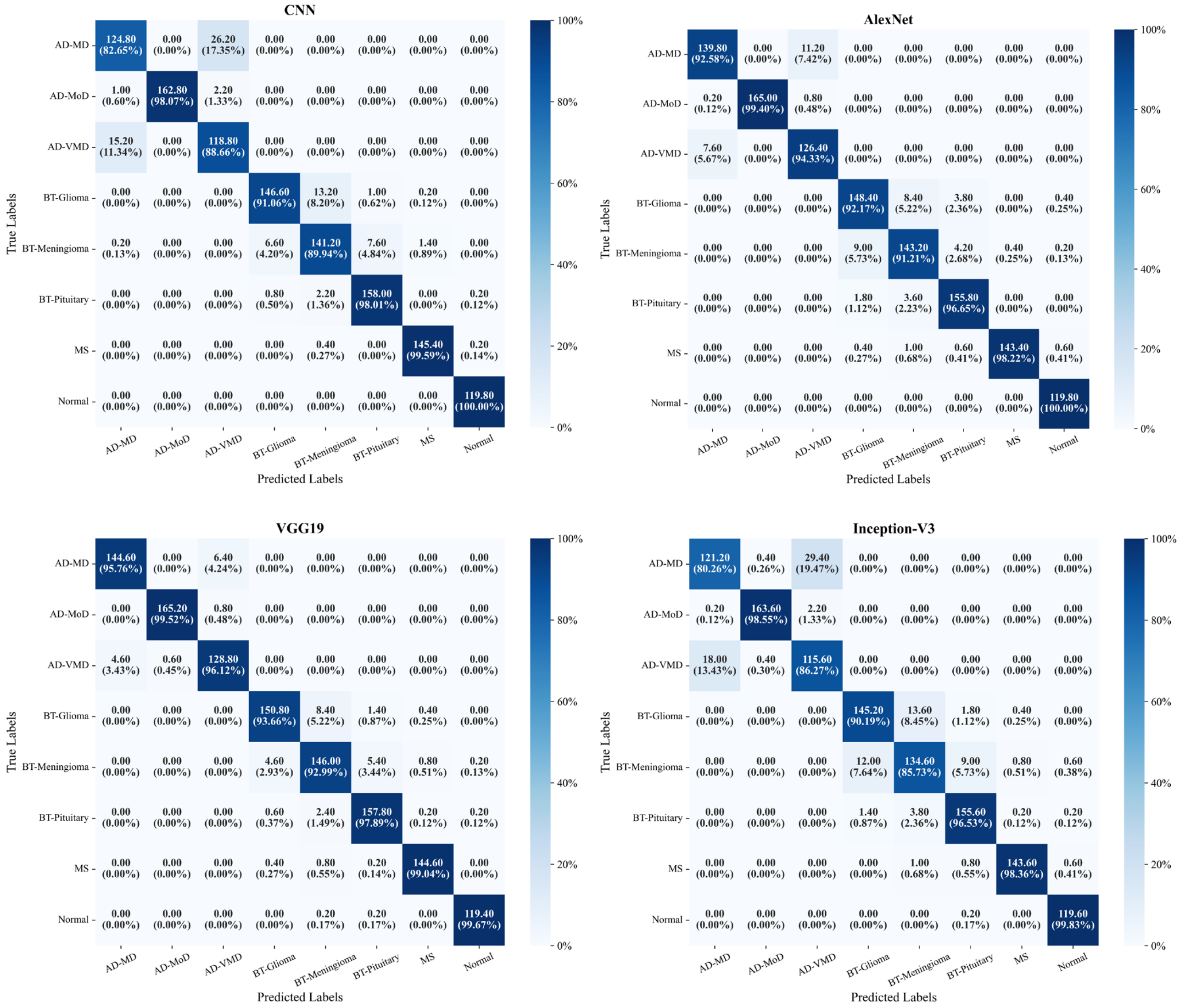

**Fig. 13** Normalized confusion matrices of the proposed End-Net and competing models

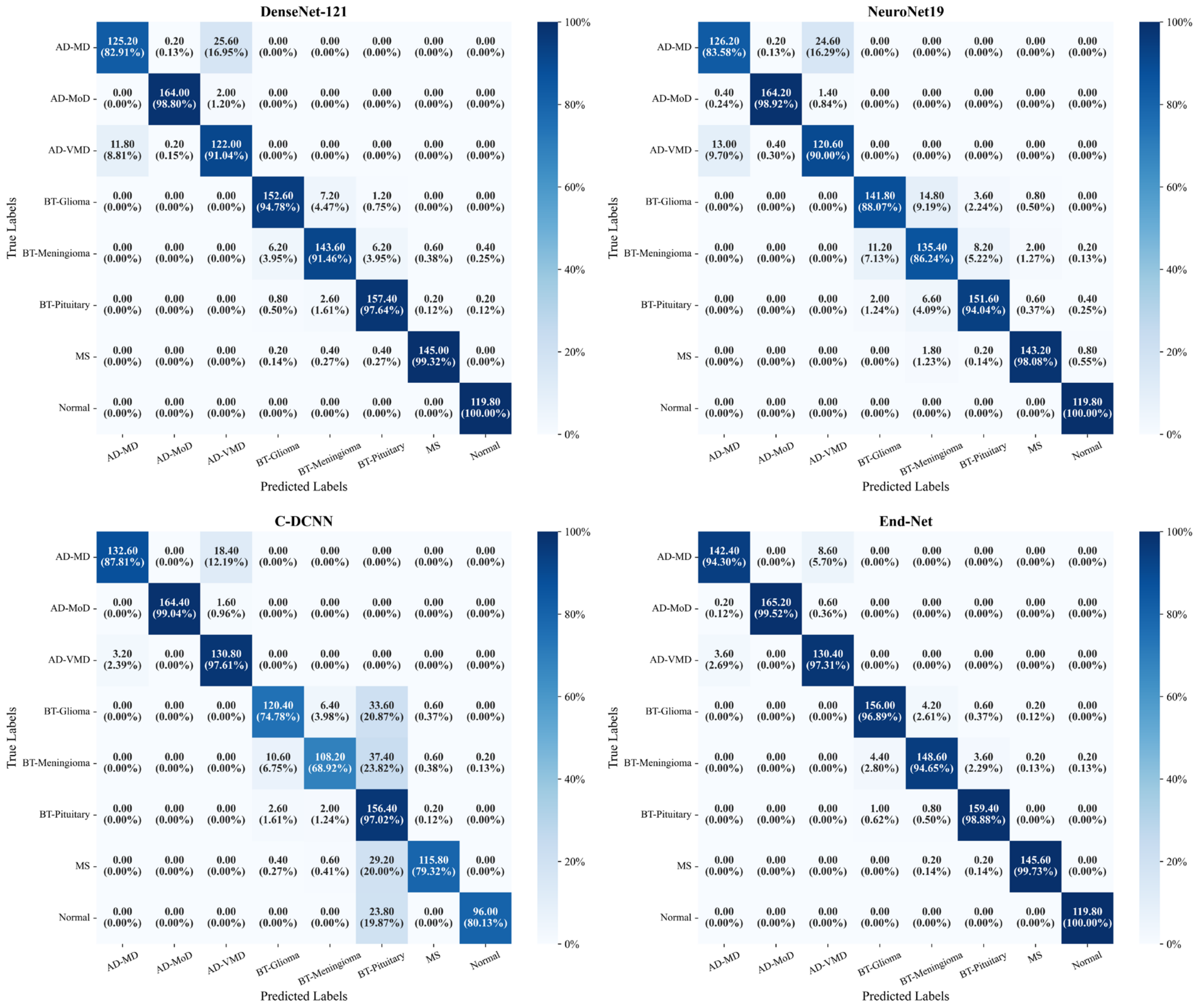

**Fig. 13** Continue

## 5. 6 Post-processing Phase and Real-Time Deployment of End-Net via Web Interface

Post-processing methods are essential for interpreting and validating deep learning model predictions by highlighting the most influential input features. In this study, the decision-making process of End-Net is explained using Gradient-weighted Class Activation Mapping (Grad-CAM), as illustrated in Fig. 14. Grad-CAM generates heatmaps that reveal the specific regions in neurological MRI data where the model focuses its attention during classification. The highlighted areas in red, orange, and yellow indicate that End-Net effectively localizes brain regions relevant to each neurological disorder. This focused attention demonstrates that the model not only achieves high classification accuracy but also relies on clinically meaningful features consistent with established neuropathology. Furthermore, End-Net's consistent ability to highlight relevant regions across cases validates its robustness and generalizability—critical factors for the real-world deployment of diagnostics.

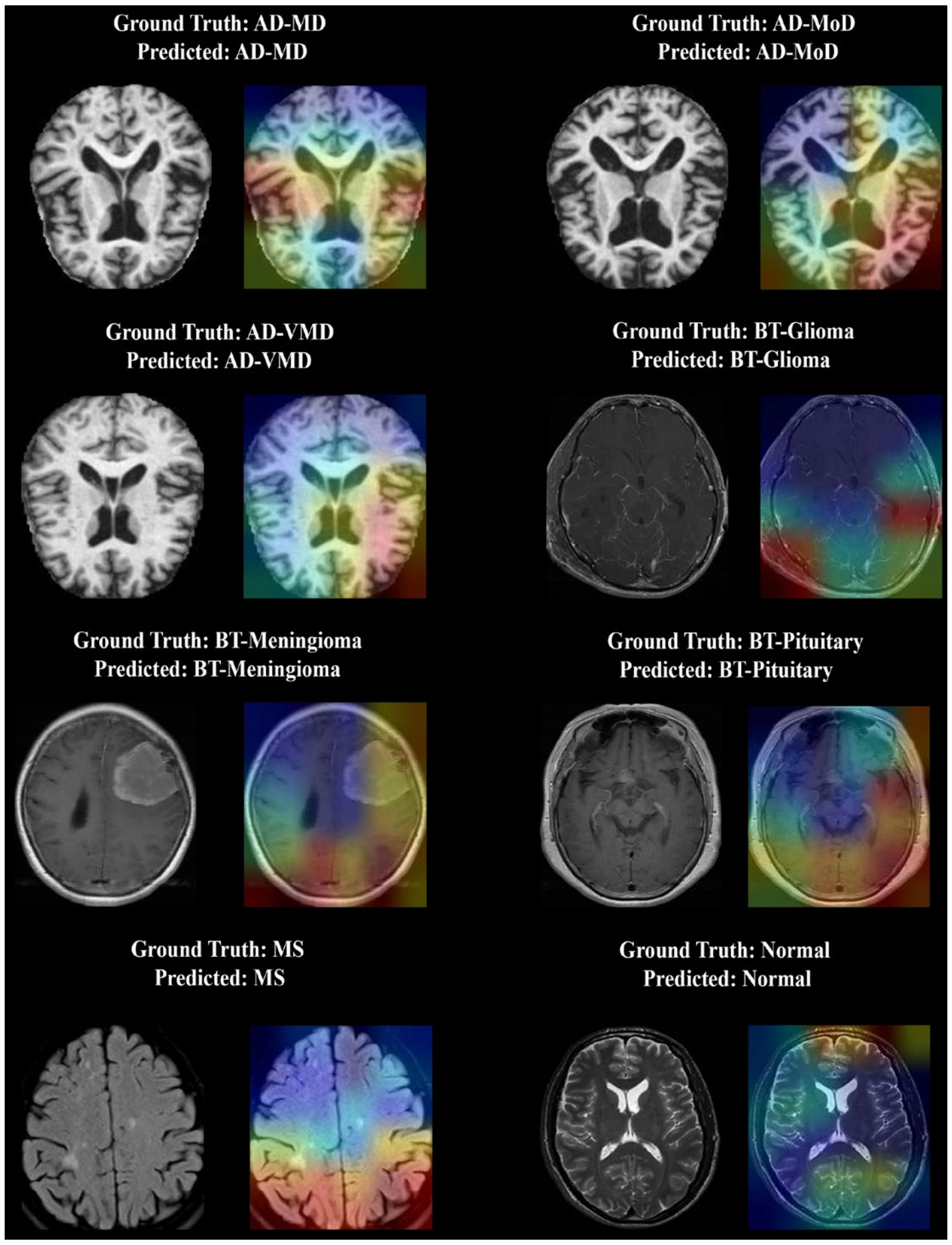

**Fig. 14** Decision-making process of the End-Net in identifying regions using Grad-CAM

Fig. 15 illustrates the proposed real-time End-Net web-based platform, accessible at https://end-net.aifatahi.space, which classifies neurological disorders into multiple categories. This integration demonstrates the successful deployment of End-Net within an online system, enabling real-time diagnostics and highlighting the model's efficiency, practicality, and clinical potential.

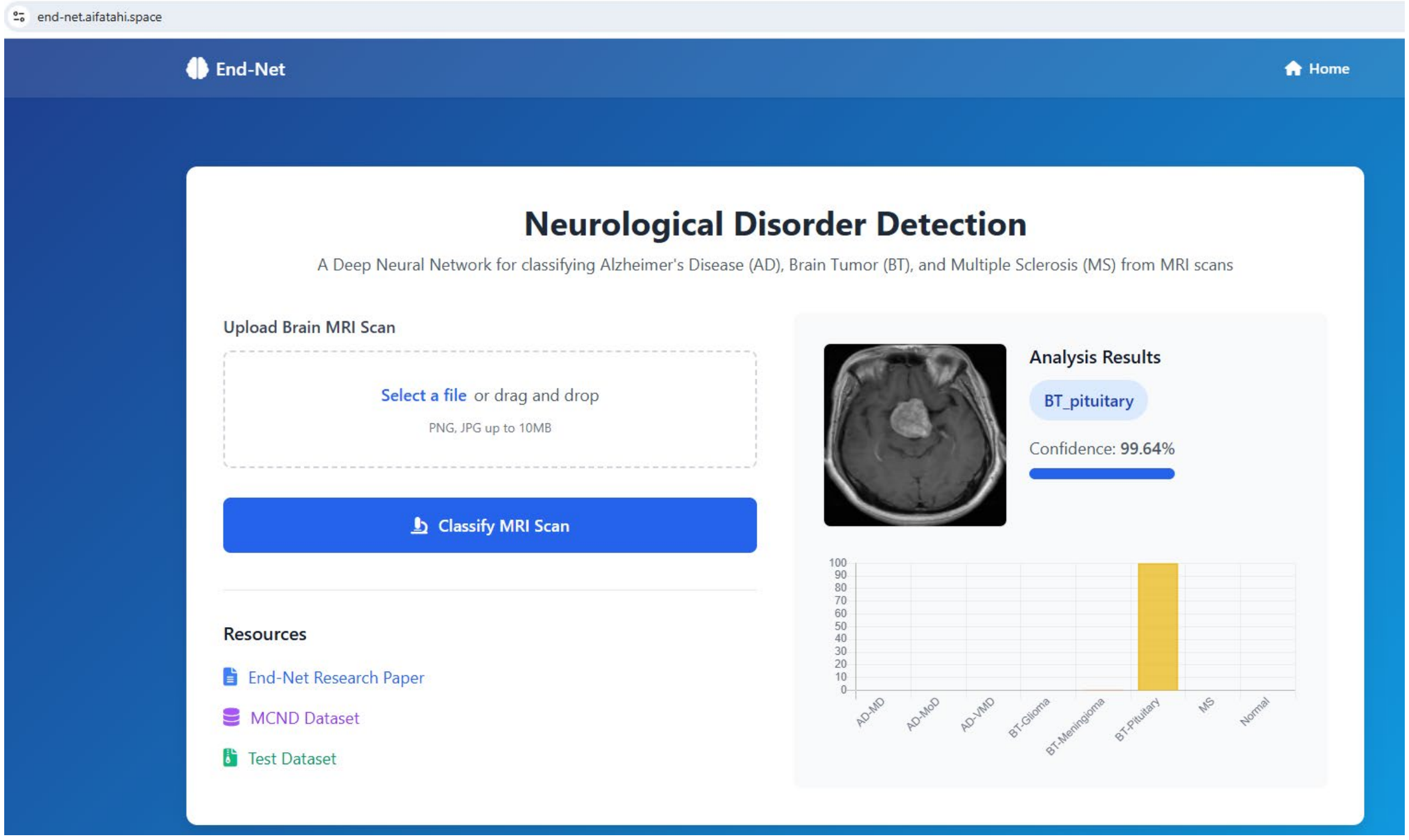

**Fig. 15** The proposed real-time End-Net web-based platform for detecting neurological disorders

## 6. Conclusion and Future Works

Neurological disorders such as Alzheimer's disease, multiple sclerosis, and brain tumors impact the peripheral nerves, spinal cord, and brain. Early detection is crucial as it can significantly enhance the five-year survival rate. Although many deep CNN models have been developed for classifying neurological disorders via MRI scans, these models often concentrate on binary classification. They may not effectively capture the multi-class features necessary to identify subtle anatomical changes across various conditions. Furthermore, they tend to lose performance when classifying multiple classes. This study presents End-Net for the multi-class classification of neurological disorders. The proposed End-Net comprises 24 convolutional layers: a three-layer convolutional stem followed by 21 optimized Inception modules engineered for efficient multiscale feature extraction. Each module employs parallel 1×1, 3×3, and factorized 5×5 convolutional paths, along with a pooled-context (max-pool) branch, enabling the network to capture fine- and coarse-grained information simultaneously. These parallel pathways jointly encode complementary cues (texture, edges, shape, and spatial context), facilitating separation of subtle inter-class anatomical differences (e.g., Alzheimer's disease vs. healthy controls) while also recognizing highly distinct phenotypes (e.g., brain tumors vs. multiple sclerosis) within a unified multi-class framework. A global average pooling head followed by a compact fully connected classifier with dropout reduces parameters and mitigates overfitting, thereby improving robustness and generalization across diverse MRI datasets.

Two sets of experiments were conducted to evaluate the proposed End-Net model against seven established deep learning models, including CNN, AlexNet, VGG19, Inception-V3, DenseNet-121, NeuroNet19, and C-DCNN. In the first set, End-Net was tested on separate neurological disorder datasets for Alzheimer's disease, brain tumors, and multiple sclerosis. Table 4 shows that for Alzheimer's disease, End-Net achieves the highest accuracy, precision, recall, and F1 score at 0.9850, with the lowest cross-entropy loss of 0.0625 and MSE of 0.0068, indicating strong and reliable classification. Table 5 demonstrates similar superiority for brain tumor classification, with an accuracy of 0.9743, precision of 0.9758, recall of 0.9743, F1 score of 0.9750, cross-entropy loss of 0.0835, and MSE of 0.0095. Table 6 presents results for multiple sclerosis, where End-Net achieves an accuracy, precision, recall, and F1 score of 0.9360, with the lowest MSE of 0.0522, despite a slightly higher cross-entropy loss of 0.2611, indicating balanced and robust performance. In the second set of experiments, End-Net's generalization was tested on the MCND dataset, which contains multiple neurological disorders. The dataset initially exhibited severe class imbalance, with the Alzheimer's disease subsets (AD-MOD, AD-MD, and AD-VMD) showing highly uneven sample counts. To mitigate bias toward the majority class, the minority classes were augmented using a WGAN-GP, while the majority class was undersampled randomly. This process balanced all Alzheimer's classes to 1,500 samples each, ensuring a uniform distribution and more stable, generalizable model training. Table 7 shows that End-Net outperforms all other models, achieving an accuracy of 0.9761, precision of 0.9766, recall of 0.9761, and F1 score of 0.9763. It also achieves the lowest cross-entropy loss of 0.1024 and the lowest MSE of 0.0050, highlighting strong diagnostic reliability and effective learning across diverse disorders.

Overall, End-Net consistently demonstrates superior performance across both domain-specific and multi-disorder datasets, providing accurate, balanced, and reliable predictions compared to other state-of-the-art deep learning models. Moreover, the End-Net's decision-making process, as visualized in Fig. 14, reveals that the proposed model identifies specific regions of the neurological MRI data that are most influential in its predictions. Despite some challenges, the End-Net showcases robustness and superior performance on the MCND dataset. Moreover, Fig. 15 illustrates the proposed real-time End-Net web-based platform for detecting neurological disorders, showcasing its real-time diagnostic capabilities and practical applications. Future research could focus on expanding the application of End-Net to classify a wider variety of neurological disorders, including epilepsy, Parkinson's disease, and Huntington's disease. Additionally, incorporating multimodal neurological data, including positron emission tomography scans, electroencephalograms, clinical biomarkers, and MRI scans, could significantly enhance the robustness of diagnosing neurological disorders with different data types.

**CRediT Authorship Contribution Statement**

**Ali Fatahi:** Writing – Original Draft, Visualization, Software, Resources, Methodology, Formal analysis, Data curation, Conceptualization, Investigation. **Hoda Zamani:** Writing - Original Draft, Visualization, Resources, Validation, Conceptualization, Formal analysis, Data curation, Project administration. **Mohammad H. Nadimi-Shahraki:** Writing – review & editing, Conceptualization, Supervision.

**Data Availability**

Both the preprocessed data and the initial data, without preprocessing, are available from the Multi-Class Neurological Disorder (MCND) dataset on Kaggle: https://www.kaggle.com/datasets/alifatahi/multi-class-neurological-disorder-mcnd-dataset/data. Upon acceptance of the manuscript, the code will be made publicly available through https://end-net.aifatahi.space.

**Declaration of Interests**

The authors declare that they have no known competing financial interests or personal relationships that could have influenced the work reported in this paper.

**Funding Declaration**

This research did not receive any specific grant from funding agencies in the public, commercial, or not-for-profit sectors.

**Ethics Statement**

This study was conducted in accordance with the highest ethical standards. The following principles guided the research.

**Data Usage and Confidentiality**

No personally identifiable information was accessed or utilized. The data were handled in accordance with data protection regulations to ensure confidentiality and integrity.

**Research Integrity**

This research's design, implementation, and reporting adhered to ethical guidelines to ensure accuracy, transparency, and reproducibility. No deceptive or misleading practices were employed in the analysis or presentation of results.